\documentclass[10pt,twocolumn,letterpaper]{article}

\usepackage[pagenumbers]{cvpr}

\usepackage{amsmath}
\usepackage{amssymb}
\usepackage{algorithm}
\usepackage{algorithmic}
\usepackage{graphicx}
\usepackage{caption}
\usepackage{booktabs}
\usepackage{multirow}
\usepackage{pifont}
\usepackage[accsupp]{axessibility}

\definecolor{turquoise}{cmyk}{0.65,0,0.1,0.3}

\newcommand{\datasetname}{3DFRONT-NC\xspace}
\newcommand{\methodname}{\textsc{CasaGPT}\xspace}

\newcommand{\hang}[1]{{\color{black}#1}}

\newcommand{\one}{\raisebox{-0.6mm}{\large{\ding{182}}}}
\newcommand{\two}{\raisebox{-0.6mm}{\large{\ding{183}}}}
\newcommand{\three}{\raisebox{-0.6mm}{\large{\ding{184}}}}
\newcommand{\four}{\raisebox{-0.6mm}{\large{\ding{185}}}}
\newcommand{\five}{\raisebox{-0.6mm}{\large{\ding{186}}}}
\newcommand{\six}{\raisebox{-0.6mm}{\large{\ding{187}}}}

\makeatletter
\let\@oldmaketitle\@maketitle
\renewcommand{\@maketitle}{\@oldmaketitle
     \vspace{-2em}
     \centering
     \includegraphics[width=\linewidth]{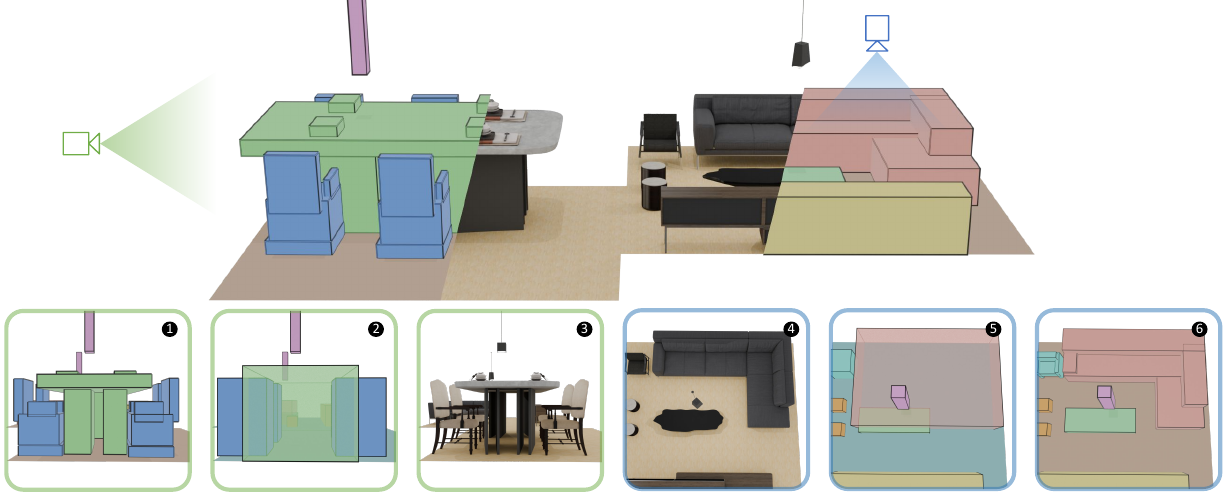}
     \captionof{figure}{We present \emph{\methodname}, an autoregressive \hang{interior design} model that frames cuboid primitives decomposed from 3D shapes into an ordered sequence for intersection-free indoor scene generation. We visualize the regional scene arrangement in \three,\four, and our model is capable of generating compact scene arrangements (\one,\six) despite bounding box intersections (\two,\five).}
     \label{fig:teaser}
    \bigskip}
\makeatother

\definecolor{cvprblue}{rgb}{0.21,0.49,0.74}
\usepackage[pagebackref,breaklinks,colorlinks,allcolors=cvprblue]{hyperref}

\title{\methodname: Cuboid Arrangement and Scene Assembly for Interior Design}

\author{Weitao Feng\textsuperscript{1}\qquad
Hang Zhou\textsuperscript{2,}\thanks{Corresponding author.}\qquad
Jing Liao\textsuperscript{3}\qquad
Li Cheng\textsuperscript{2}\qquad
Wenbo Zhou\textsuperscript{1}\\
\textsuperscript{\rm 1}University of Science and Technology of China\qquad 
\textsuperscript{2}University of Alberta\\
\textsuperscript{3}City University of Hong Kong\\
}

\begin{document}

\maketitle

\begin{abstract}
We present a novel approach for indoor scene synthesis, which learns to arrange decomposed cuboid primitives to represent 3D objects within a scene. Unlike conventional methods that use bounding boxes to determine the placement and scale of 3D objects, our approach leverages cuboids as a straightforward yet highly effective alternative for modeling objects. This allows for compact scene generation while minimizing object intersections. Our approach, coined \methodname for Cuboid Arrangement and Scene Assembly, employs an autoregressive model to sequentially arrange cuboids, producing physically plausible scenes. By applying rejection sampling during the fine-tuning stage to filter out scenes with object collisions, our model further reduces intersections and enhances scene quality. 
Additionally, we introduce a refined dataset, \datasetname, which eliminates significant noise presented in the original dataset, 3D-FRONT.
Extensive experiments on the 3D-FRONT dataset as well as our dataset demonstrate that our approach consistently outperforms the state-of-the-art methods, enhancing the realism of generated scenes, and providing a promising direction for 3D scene synthesis. Code is available at \url{https://github.com/CASAGPT/CASA-GPT}

\end{abstract}    
\section{Introduction}
\label{sec:intro}

Creating a comprehensive model capable of generating sophisticated and realistic furnished interior spaces for virtual walk-throughs is a challenging task in computer graphics, with wide-range applications in game development, VR/AR, and interior design. Scene synthesis also enables automatic environment creation for products displayed in retail, films, and games~\cite{zhao2021luminous}, while providing valuable data for training AI models in 3D scene understanding \cite{deitke2022️,chang2017matterport3d,cho2024language}.

Traditional indoor scene synthesis approaches framed the problem as an optimization task with predefined constraints, often guided by design principles and spatial guidelines \cite{merrell2011interactive,fisher2012example} rooted in aesthetic and ergonomic criteria~\cite{fisher2011characterizing,yu2015clutterpalette,leimer2022layoutenhancer}. These methods relied on rules tailored for modeling object relationships (e.g., table-chair pairs) and handcrafted constraints on human-object interactions (e.g., affordance maps) to guide layout generation \cite{fisher2015activity,fu2017adaptive,jiang2012learning}. However, these approaches were computationally expensive and struggled to scale for more complex scenes.

Recent approaches addressed these limitations using data-driven methods that leverage deep generative models for interior design. For instance, autoregressive models improved scene realism by sequentially predicting object properties~\cite{wang2021sceneformer, paschalidou2021atiss}, while diffusion-based models enabled high-quality scene synthesis with diverse range of outputs~\cite{tang2024diffuscene, zhai2024commonscenes}. Despite their impressive capabilities, these models often produce physically implausible scenes with significant object collisions. This raises a critical question: \textit{What causes object collisions during scene generation?} We aim to address this issue from following two perspectives:

(\textbf{i}) Training data often contain significant noises with intersecting objects (e.g. \Cref{fig:dataset_cmp}). We observe that most previous methods represent mesh objects as bounding boxes, which have inherent limitations, as box intersections often do not align with actual mesh intersections, as shown in \Cref{fig:teaser}. 
To address the challenges of modeling spatial relationships among meshes, we treat each mesh as an assembly of cuboids. These cuboids can then be effectively modeled using autoregressive approaches for scene generation.

(\textbf{ii}) Given the probabilistic nature of generative models~\cite{murphy2012machine}, object collisions are inevitable due to the inherent uncertainty in the generation process. However, the collision rate can be significantly reduced by incorporating rejection sampling \cite{liu2023statistical,yuan2023scaling,khaki2024rs} into scene synthesis, to \textit{reject} scenes with bounding box intersections during fine-tuning.

Built upon this analysis, we present a simple, effective, and compact autoregressive model, \methodname, capable of arranging scene objects without intersections. 
\methodname utilizes the cuboid primitives, decomposed from meshes, as input tokens for scene modeling. With this approach, we can achieve a $90.6\%$ non-occlusion rate for bedrooms. 

\noindent Our contributions are summarized as follows:
\begin{description}[noitemsep, parsep=3pt, leftmargin=5pt]
    \item[Cuboid-Based Scene Representation.] We model shapes using cuboid primitives for interior scene synthesis, ensuring compactness and effectively avoiding object collisions. 
    \item[Advanced Autoregressive Modeling.] \methodname employs a Llama-3 based autoregressive model for cuboid sequence modeling and integrates rejection sampling during fine-tuning to enable more refined scene generation. 
    \item[New Dataset Curation.] We introduce \datasetname, a refined dataset with minimized object intersections, where objects are represented by cuboid assemblies. 
    \item[Comprehensive Evaluation.] Extensive experiments demonstrate that \methodname significantly reduces object intersections and improves generation quality across various scenarios. 
\end{description}

\section{Related work}

\paragraph{Heuristic scene synthesis.}

A great many classical methods in computer graphics employ heuristics and predefined rules to restrict object positioning~\cite{xu2002constraint,yu2011make,weiss2018fast}.
Make-it-Home~\cite{yu2011make} investigated human factors, such as accessibility, visibility, and pathway constraints for the evaluation of synthetic scenes. 
Works have also focused on scene generation and evolution navigated by human activities~\cite{fisher2015activity, ma2016action}. 
An alternative approach is procedural modeling~\cite{qi2018human,purkait2020sg}, which recursively applies a set of functions for scene synthesis. 
SG-VAE~\cite{purkait2020sg} learned co-occurrences, shapes and poses through a grammar-based auto-encoder~\cite{kusner2017grammar}. 
In contrast, our model implicitly captures inter-object relationships without the need for hand-crafted constraints.

\begin{figure*}[t]
\centering
\includegraphics[width=1.0\linewidth]{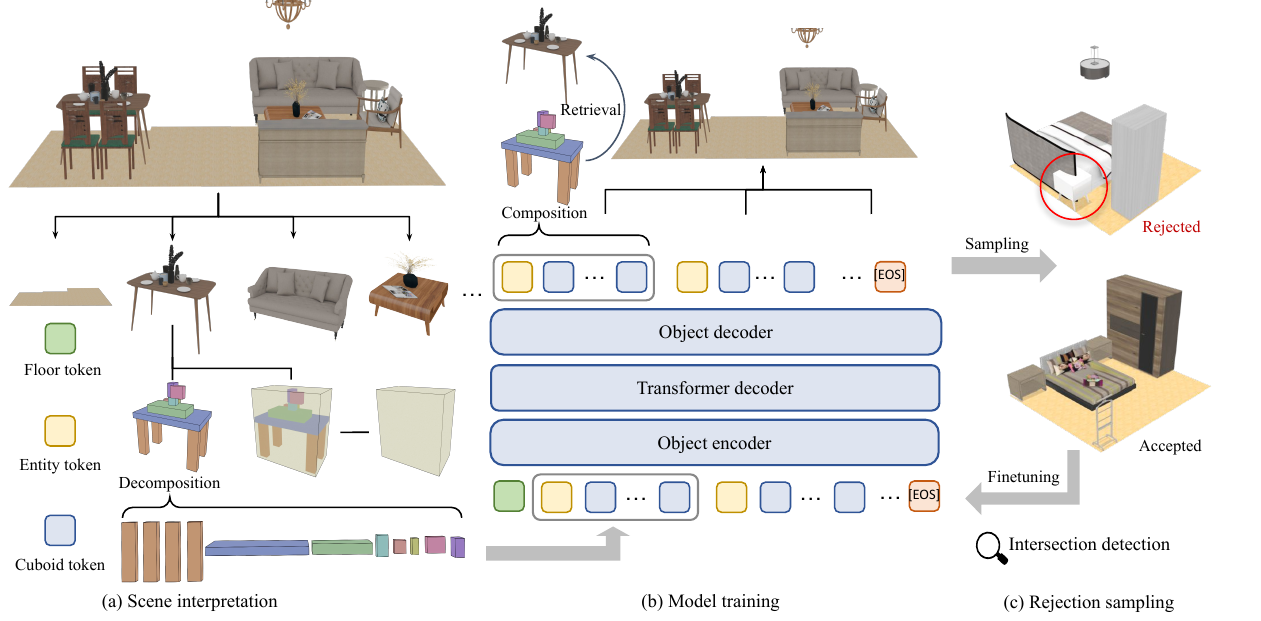}
\caption{Overview of the \methodname framework for indoor scene generation. (a) \textbf{Model Pre-training}: Objects in the scene are encoded using distinct tokens representing entities and cuboids. An autoregressive training process in the Transformer Decoder enables the generation of complex layouts. (b) \textbf{Rejection Sampling}: Iterative refinement is applied through cuboid collision inspection to reject unsuitable generation results, followed by model fine-tuning, resulting in high-quality, realistic 3D layouts.}
\label{fig:method}
\end{figure*}

\vspace{1.5em}\noindent \textbf{Graph modeling for scene synthesis.}
Representing indoor scenes as graphs has been widely explored in~\cite{fisher2012example, wang2019planit, 
li2019grains, zhou2019scenegraphnet, gao2023scenehgn, wei2023lego, tang2024diffuscene, zhai2024commonscenes}. 
Fisher et al.~\cite{fisher2012example} introduced a probabilistic model based on Bayesian networks and Gaussian mixtures that can be trained with few-shot learning. 
GRAINS~\cite{li2019grains} represented scenes as parsing trees with the location of each node annotated by its relative position to the parent node.  \textsc{SceneHGN}~\cite{gao2023scenehgn} extended this approach and captured the full hierarchy from the room level to object and object part levels. 
SceneGraphNet~\cite{zhou2019scenegraphnet} employed graph generative networks to facilitate message passing, enabling the learning of node representations and their interactions.
CommonScenes~\cite{zhai2024commonscenes} and EchoScene~\cite{zhai2025echoscene} both introduced a dual-decoder  
for graph generative models, with separate decoders dedicated to layout and shape generation. 
LEGO-Net~\cite{wei2023lego} and DiffuScene~\cite{tang2024diffuscene} employed denoising diffusion models by representing scenes as fully connected graphs and DeBaRA~\cite{maillard2024debara} extended it to score based model.
\textsc{InstructScene}~\cite{lininstructscene} incorporated pairwise spatial relations (e.g., left of, above) into graph construction 
for fine-grained scene editing. 
\textsc{PhyScene}~\cite{yang2024physcene} incorporated a collision loss between two bounding boxes to prevent object intersections. 
Despite the capabilities of various generative models, these methods still fail to generate scenes without mesh intersections or handle intersections at the mesh level.

\vspace{1.5em}\noindent \textbf{Sequence models for scene synthesis.}
Pioneering efforts on autoregressive indoor scene synthesis models~\cite{wang2021sceneformer,paschalidou2021atiss,sun2024forest2seq,cofs} operate on sequences of scene nodes, allowing transformer attentions for enhanced contextual understanding. 
SceneFormer~\cite{wang2021sceneformer} employs separate transformers to model each object attribute, with objects ordered based on class frequency. 
Concurrently, ATISS~\cite{paschalidou2021atiss} proposed an order-invariant transformer by introducing Monte Carlo permutation into order construction and removing positional encoding. 
COFS~\cite{cofs} incorporated the encoder-decoder architecture from BART~\cite{lewis2020bart} into scene synthesis, allowing for bidirectional attention over scene objects. 
\textsc{Forest2Seq}~\cite{sun2024forest2seq}
organized object orders based on scene hierarchy, enabling efficient semantic sequence-to-sequence learning. 
LayoutGPT~\cite{feng2023layoutgpt} introduces a training-free method that infuses visual commonsense into large language models, enabling text-guided scene generation. 
These efforts have primarily focused on order-sensitive sequence learning, but have yet to demonstrate the ability to mitigate object intersections. 
In contrast, our approach learns the arrangement of decomposed scene primitives such as cuboids, which enhances the compactness of arrangement and avoids intersection.

\vspace{1.5em}\noindent \textbf{Shape abstraction. }
Extensive research has explored different forms of 3D shape representation, such as cuboids~\cite{tulsiani2017learning}, convexes~\cite{deng2020cvxnet} and superquadrics~\cite{paschalidou2019superquadrics}. 
Cuboid abstractions, known for their simplicity, have recently been influenced by shape segmentation~\cite{yang2021unsupervised}.
Recent approaches, such as DPF-Net~\cite{shuai2023dpf}, used deformable primitives to progressively reconstruct shapes, enabling high-quality detail recovery and consistent structure.
ShapeAssembly~\cite{jones2020shapeassembly} leverages procedural programs to generate complex 3D structures from parameterized cuboids, allowing for both interpretability and flexibility. In contrast, our cuboid abstraction process does not rely on any pretrained domain-specific models, as it is designed to generalize across various meshes.

\section{\methodname}

We introduce \methodname, an autoregressive scene generation model customized for minimizing intersection occurrences among objects. First, we built fine-grained cuboid primitives for each piece of furniture and decomposed the entire scene into a cuboid assembly. Our goal is to directly model these sets of primitives (see \Cref{fig:method}) using an autoregressive model. We then further enhance the model's ability to avoid intersections through rejection sampling. Finally, using this fine-grained shape representation, we constructed the \datasetname dataset, which effectively reduces object intersection, further refining the training data.

\subsection{Generate cuboid shape abstraction}

\vspace{0.5em}\noindent \textbf{Voxelization.} The first step in our approach involves normalizing a given 3D mesh to fit within a fixed grid size of $N^3$ with $N$ being the voxel resolution. We convert the 3D mesh into a voxelized form by calculating the occupancy of each grid cell, the mesh's internal space is also filled. The result is a voxel grid where each cell represents whether it is occupied, leading to a binary representation $O \in \mathbb{R}^{N^3}$.

\begin{figure}[t]
\centering
\includegraphics[width=0.8\linewidth]{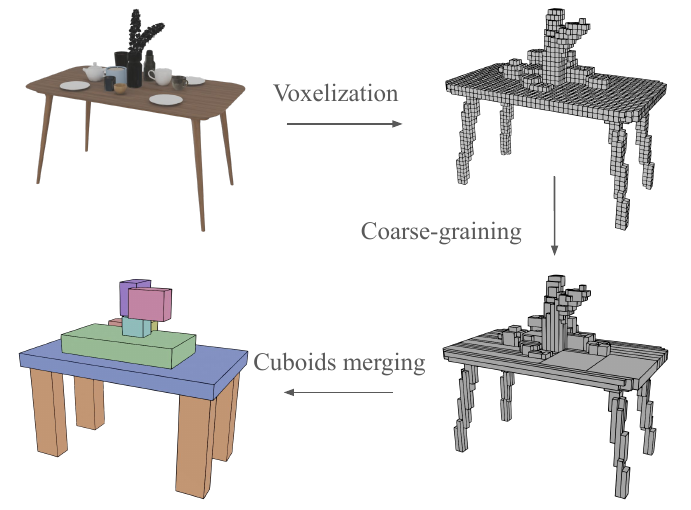}
\caption{Workflow of the voxelization, coarse-graining, and cuboid merging process, transforming a 3D object into a compact cuboid representation.}
\label{fig:dataset_flowchart}
\vspace{-0.5em}
\end{figure}

\vspace{0.5em}\noindent \textbf{Voxel grid coarse-graining.}
To simplify the geometric representation, we coarsen the voxel grid into larger cuboidal structures, reducing the number of discrete voxels \textit{without increasing the volume occupied by the objects}. Starting from each unassigned occupied voxel, we expand along the $x$- and $z$-axes to form maximal contiguous cuboids, marking voxels as assigned after incorporation. This process continues until all occupied voxels are merged, preserving structure while reducing discrete elements. The result of this can be seen in the bottom-right corner of \Cref{fig:dataset_flowchart}.

\vspace{0.5em}\noindent \textbf{Merging cuboids.}
After the initial coarse-graining step, where cuboids were formed by merging contiguous voxels, we further reduced the number of cuboidal structures, simplifying the geometric representation.
Based on the cuboid sequence obtained in the previous step, we iteratively select pairs of cuboids and assess their potential for merging. Specifically, given two cuboids, \( A \) and \( B \), we calculate the volume of a new cuboid \( C \), which is the bounding cuboid containing both \( A \) and \( B \). The new cuboid \( C \) is only accepted as a valid merged structure if \( \frac{V_C}{V_A + V_B} < \tau \), where \( V_C \) is the volume of the merged cuboid, \( V_A \) and \( V_B \) are the volumes of the original cuboids, and \( \tau \) is the threshold. If the condition is met, the cuboids are merged into a new cuboid \( C \); otherwise, they remain separate.

Building on this, we found that performing an initial simple segmentation on the target voxel \( O \), followed by cuboid merging within each split before conducting a global merge, further improves the results. We first project the voxels onto a plane and then use bounding boxes to outline the target object. By incrementally removing empty regions (areas with a value of 0) within the rectangles, we refine a coarse bounding rectangle into multiple tighter rectangles that more closely fit the target shape, thereby minimizing the total coverage area. These rectangles are then used to segment \( O \). 
In addition, we introduced a dynamic threshold mechanism to encourage the merging of smaller cuboids while penalizing the merging of larger ones. For details, refer to the supplementary material.

\subsection{Scene representation modeling}
In our approach, a scene \(\mathcal{S}\) is defined as \(\{\textbf{F}, o_i\}_{i=1}^N\), where \(\textbf{F}\) represents the floor plane and up to \( N \) objects are included as \(\{o_i\}_{i=1}^N\). Each object is encoded with an entity token followed by several cuboid tokens sorted in ascending order of their heights; see \Cref{fig:method}. This tokenization scheme encodes critical attributes: object class \( c \in \mathbb{R}^C \), size \( s \in \mathbb{R}^3 \), translation \( t \in \mathbb{R}^3 \), and rotation angle \( \theta \in \mathbb{R} \) around the vertical axis. For angle representation, we use the SO(3) representation from \cite{yin2021center}, encoding the rotation angle as \( \sin \theta \) and \( \cos \theta \) to enhance continuity compared to a simple scalar angle representation. In summary, each object \( o_i \) is represented by a concatenation of all attributes:

\begin{equation}
o_i = [ c_i, t_i, s_i, \sin \theta_i, \cos \theta_i] \in \mathbb{R}^D,
\label{eq:2}
\end{equation}
where \( D \) is the dimension of the combined attributes.

\vspace{0.5em}\noindent \textbf{Sequence representation.}
We represent a scene as a sequence. The \textit{entity token} denotes an object's overall position, including translation, size, and rotation. To differentiate each object in the sequence, we assign the entity token a unique class label, \(\texttt{[SEP]}\), which serves as a delimiter. The \textit{cuboid tokens} represent the cuboids that comprise the object, each holding attributes such as translation, size, rotation, and a class label indicating the object’s type. 

To construct a scene, we start with the floor plane token \( \textbf{F} \), followed by all furniture tokens within the scene. During training, the order of the furniture items is randomly permuted, while the cuboids within each piece of furniture are sorted in ascending order of their heights.

\subsection{Rejection sampling}

Rejection sampling is a Monte Carlo method to generate samples from a complex distribution by first sampling from a simpler one and then \textit{accepting} or \textit{rejecting} those samples based on certain criteria.
It can be effectively applied in scene synthesis to enhance spatial coherence and realism by minimizing object intersection. Specifically, it iteratively refines generated layouts, excluding samples with object collisions or unrealistic arrangements, improving adherence to physical constraints.

In our approach, we construct an approximation of an optimal scene layout policy \(\pi^*\) by applying rejection sampling to candidate layouts generated from a base policy \(\pi\). We first generate \(K\) candidate samples \((S_1, \ldots, S_K)\). Each scene sample is then evaluated by calculating the 3D IoU of the generated cuboids, rejecting layouts that result in object intersection by a predefined threshold \(T\). The filtered samples are then used to continue fine-tuning our model.

This process of rejection sampling can be iteratively applied. In each iteration, an initial set of rollouts \(\textbf{S}_1\) is created, filtered to remove suboptimal results, and then distilled back into the scene generation policy \(\pi_i\) via fine-tuning. Each subsequent policy \(\pi_{i+1}\) is fine-tuned based on the distilled dataset \(\textbf{S}_i = \textbf{R}_i \cup \textbf{S}_{i-1}\), where \(\textbf{R}_i\) represents new, high-quality samples generated by \(\pi_i\). By iterating this procedure, the model converges towards generating scenes that consistently respect spatial constraints, reducing intersection and increasing layout plausibility.

\subsection{Training and inference}
\noindent \textbf{Object retrieval.}
During inference time, we select 3D models from the 3D-FUTURE dataset \cite{fu20213d} and place them in the scene. Our model can generate fine-grained cuboid representations, so we perform retrieval by calculating the IoU match between the generated results and models in the database. Specifically, we first filter out objects in the 3D-FUTURE dataset with the same class label \(c\) as the generated object, forming the subset \(\textbf{S}\{c\}\). Next, we define a voxel space \(V \in \mathbb{R}^{N^3}\) and compute the occupation grid for the cuboids of the normalized generated model. Then, we compute the 3D IoU in \(V\) between the generated model and all models in \(\textbf{S}\{c\}\), selecting the one with the highest IoU.

\begin{figure}[t]
\centering
\includegraphics[width=0.95\linewidth]{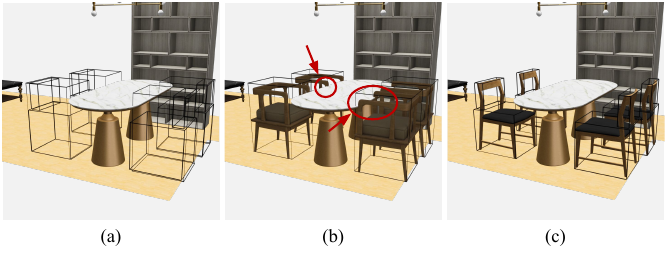}
\caption{Comparison of object retrieval methods: (a) Layout model predictions with bounding boxes and cuboids. (b) Nearest neighbor results using bounding boxes, causing intersection issues. (c) Object retrieval using our proposed method prevents intersection, enhancing retrieval accuracy.}
\label{fig:obj_ret}
\vspace{-0.5em}
\end{figure}

\Cref{fig:obj_ret} demonstrates the advantages of our proposed object retrieval method. Based on the similarity of bounding boxes can lead to the selection of objects that cause the collision. However, using a fine-grained cuboid representation for retrieval avoids intersection.

\vspace{0.5em}\noindent \textbf{Model architecture.}
We use Llama-3~\cite{dubey2024llama} model structure as our backbone transformer decoder with RMSNorm~\cite{zhang2019root}, SwiGLU~\cite{shazeer2020glu}, and rotary position encoding (RoPE)~\cite{su2024roformer}. 
However, in our task, we found RoPE less effective than learning-based position encoding, likely due to the relatively short sequence length, where RoPE tends to underperform. Therefore, we replaced the position encoding with learning-based position encoding. 

\vspace{0.5em}\noindent \textbf{Training objectives.}
Following ATISS~\cite{paschalidou2021atiss} we use its attribute extractor as the object decoder to extract the content of tokens. Cross-entropy loss is used for the predicted label \( c \) and the ground truth label \( c_{\text{tr}} \). We model size \(s\), translation \(t\), and angle \( [\sin \theta, \cos \theta] \) with a mixture of logistic distributions as \( x_j \sim \sum_{k=1}^K \pi_k^x \text{Logistic}(\mu_k^x, \sigma_k^x) \), where \( x \in \{s, t, r\} \), and \( \pi_k^x \), \( \mu_k^x \), and \( \sigma_k^x \) represent the weight, mean, and variance of the \(k\)-th logistic distribution for each attribute \(x\). Then, we compute the loss between the label token and the predicted token using negative log-likelihood (NLL).
We use teacher forcing for training. In contrast to ATISS, which trains by predicting one token at a time, teacher forcing improves training efficiency and reduces training time.

\subsection{Dataset curation}
\label{sec:dataset}

\begin{figure}[h!]
\centering
\includegraphics[width=0.95\linewidth]{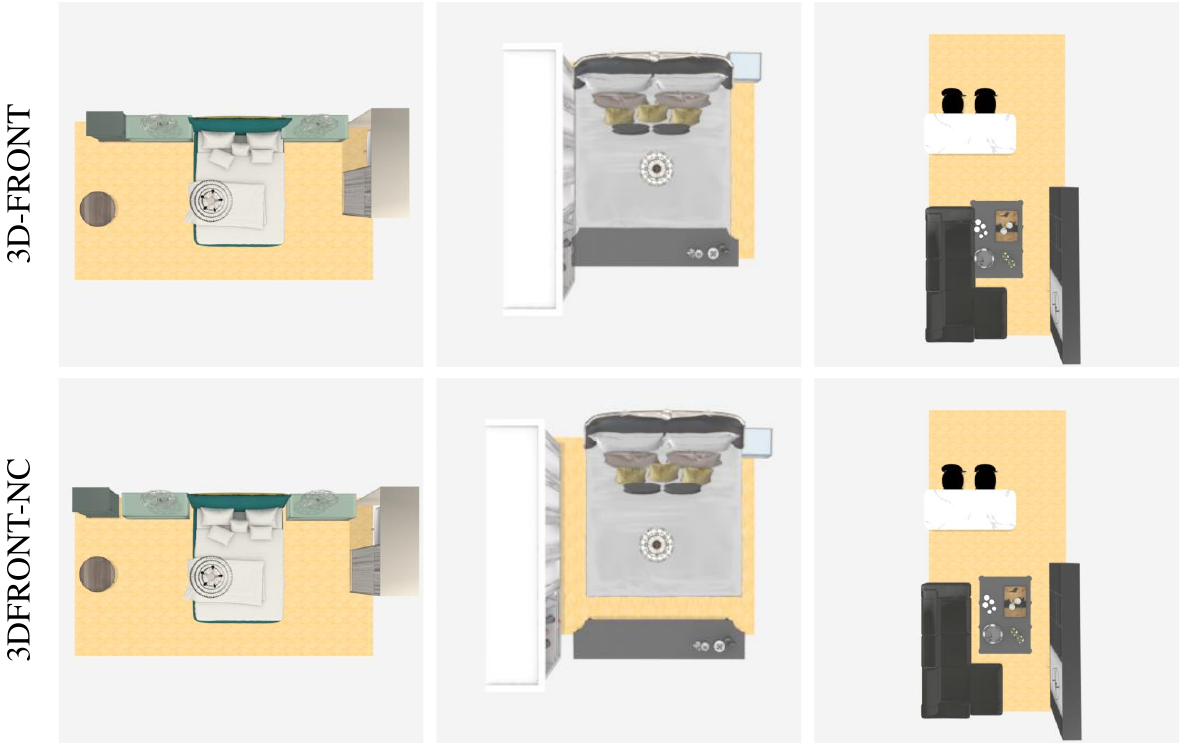}
\caption{Comparison of the dataset before and after applying intersection avoidance. Our method effectively adjusts object positions to prevent intersections while preserving non-intersecting parts of the scene (third column).}
\label{fig:dataset_cmp}
\vspace{-0.5em}
\end{figure}

Learning-based methods directly reflect the data distribution of the dataset itself. Unfortunately, the 3D-FRONT dataset contains numerous instances of object intersection. Based on the fine-grained shape representation, we can effectively distinguish between bounding box overlap and object intersection to refine the training data further. We propose the \datasetname dataset, where NC stands for ``noise clean'', significantly reduces object intersection in the data.

We begin by calculating the Intersection-over-Union (IoU) matrix for cuboids~\cite{zhou2019iou} in the scene, excluding intersections between cuboids belonging to the same entity. Next, we implement an iterative avoidance mechanism to adjust cuboid positions. The translation update is expressed as \(\mathbf{t}_{\text{new}} = \mathbf{t}_{\text{old}} - \eta \cdot \text{clip} (\nabla \sum M_{\text{IoU}}) \), where $M_{\text{IoU}}$ is the IoU matrix, $\eta$ is the optimization rate and $\text{clip}$ is gradient clipping. These updates are carried out using gradients derived from the loss function. This iterative process continues until convergence is achieved or reaches a predefined maximum number of iterations. We refrain from updating the translations along the y-axis, as such adjustments lead to objects within the scene floating.

\Cref{fig:dataset_cmp} illustrates dataset adjustments before and after applying intersection avoidance. Our method performs subtle adjustments to the objects in the scene to avoid object intersections. For objects that initially did not have intersections (the table and chair combination in the third column), our method does not make any unnecessary adjustments.

\section{Experiment}

\begin{figure*}[t]
\centering
\includegraphics[width=0.95\linewidth]{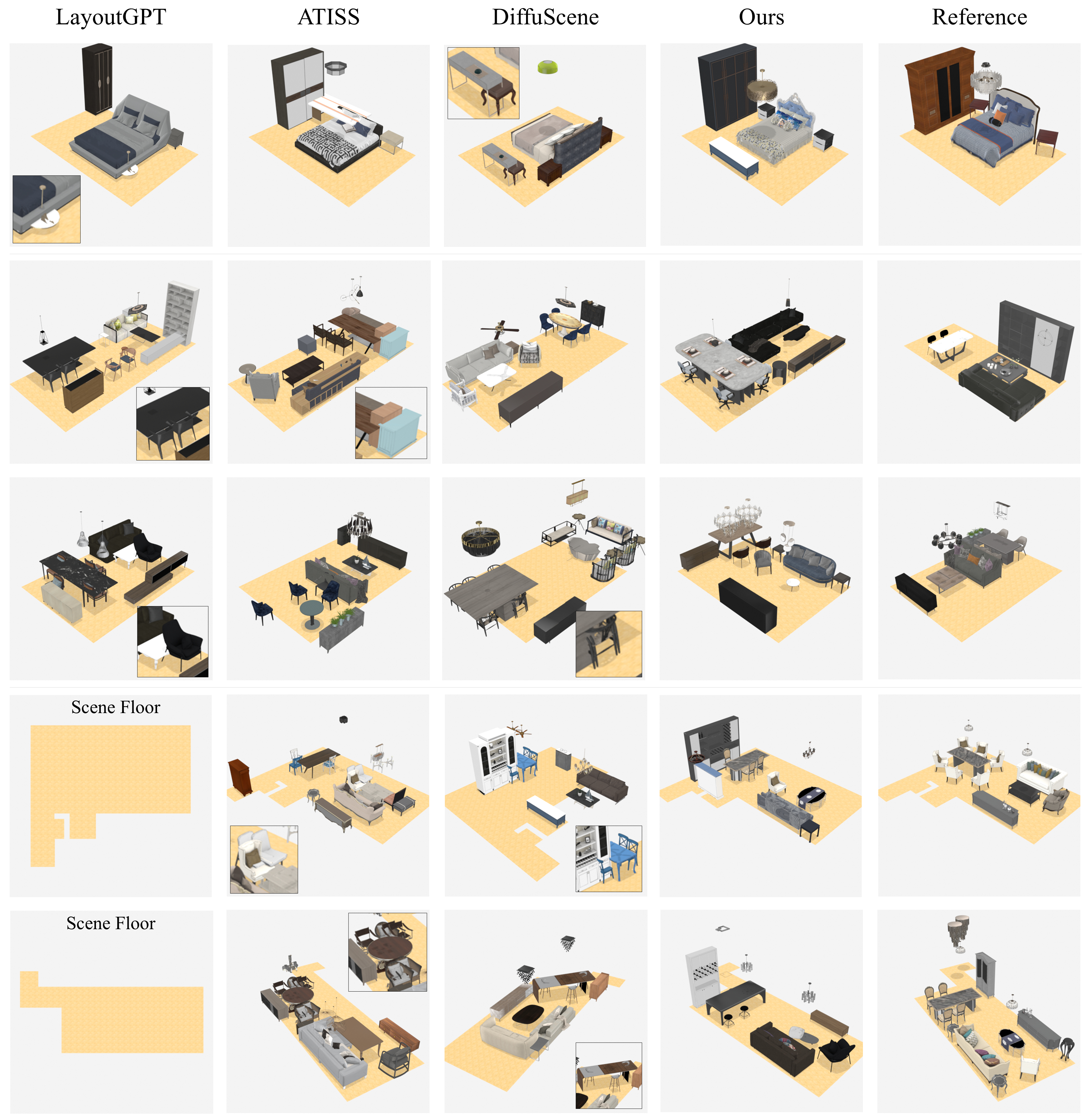}
\caption{We provide a qualitative comparison between our method, previous methods, and references from \datasetname. The first row shows \textit{bedroom} scenes, the second and third show \textit{living/dining rooms}, and the fourth and fifth depict \textit{living/dining rooms} with complex floor plans, unsupported by LayoutGPT and DiffuScene. Our method demonstrates superior generation results and better object intersection avoidance across various scenarios. We also provide zoomed-in views to highlight intersections.
}
\vspace{-0.5em}
\label{fig:main_cmp}
\end{figure*}

\noindent \textbf{Datasets.}
We used the 3D-FRONT dataset~\cite{fu20213front}, which contains 6,813 residential environments and 14,629 rooms, as our training and testing data. The rooms are furnished with high-fidelity 3D models from the 3D-FUTURE dataset~\cite{fu20213d}. Following the ATISS methodology, we focused on four room categories: 4,041 bedrooms, 900 dining rooms, 813 living rooms, and 286 libraries. In our refined \datasetname dataset, we retained the original number of scenes. For the cuboid abstraction process, we set the voxel resolution to $N = 64$.

\vspace{0.5em}\noindent \textbf{Implementation details.}
We trained our model on both the 3D-FRONT and \datasetname datasets using an NVIDIA A6000 GPU and the AdamW optimizer~\cite{loshchilov2017decoupled} with a learning rate of 0.0001. The training lasted 1500 epochs with a cosine scheduler and a 10\% warm-up. We sample 15,000-20,000 instances during rejection sampling and set the threshold \( T \) when the average cuboid IoU fell below 0.001, ensuring at least 10,000 samples succeed. We then trained for 1,000 epochs and repeated the process three times. All models were trained with a batch size of 128. Model architecture details are provided in the supplementary material. We encode the floor plan \( \mathbf{F} \) using a ResNet18 network into a 64-dimensional vector, then map it to the hidden dimension via a linear layer as the start token.

\vspace{0.5em}\noindent \textbf{Evaluation metrics.}
Following previous works \cite{paschalidou2021atiss,wang2021sceneformer,tang2024diffuscene}, we use Fréchet Inception Distance (FID)~\cite{heusel2017gans} to evaluate the quality and diversity of the synthesized layout images, and Category KL Divergence (CKL × 0.01) to measure the plausibility and diversity of 1,000 synthesized scenes. We propose two new metrics, Cuboids Intersection-over-Union (CIoU) and non-intersection rate (NIRate), to measure the collision in generated results.

For FID, we render generated and ground-truth scenes as images with 256×256 resolutions through top-down orthographic projections, with the same floor textures across paired samples. For CIoU, we calculate the volume for intersections between fine-grained cuboids in 3D space and divide it by the volume of all cuboids in the scene. We scale them by 1,000 for better clarity. Note that we do not calculate intersections between cuboids within individual objects. For methods that do not directly generate cuboid-based shapes, we retrieve their cuboid-based shapes by the index in 3D-FUTURE from the generated scene. 
NIRate reflects the proportion of high-quality scenes with minimal intersections. We calculate it by first filtering CIoU results with a threshold of 0.01, then obtaining the pass rate among 1,000 sampled scenes.

\subsection{Scene generation}
\noindent \textbf{Qualitative results.}
We present qualitative evaluations in \Cref{fig:main_cmp}. LayoutGPT~\cite{feng2023layoutgpt}, ATISS~\cite{paschalidou2021atiss}, and DiffuScene~\cite{tang2024diffuscene} are more likely to produce object intersections and unfavorable scene layouts. The references come from \datasetname, and our results are trained on our own dataset.
Our method effectively differentiates between bounding box overlap and actual object intersection, significantly reducing the latter. For example, the bounding boxes of the L-shaped sofa and coffee table overlap without object intersection, as do those of the dining chair and dining table.

\vspace{0.5em}\noindent \textbf{Quantitative results.}
\Cref{tab:main results} presents the quantitative comparisons under various evaluation metrics. Our method outperforms others in FID, CIoU, and NIRate, particularly showing a 31\% advantage in NIRate compared to previous methods. While DiffuScene achieves comparatively higher performance on the CKL metric, we believe this comparison is not entirely apples-to-apples, as DiffuScene does not consider the floor plane in its evaluation, thereby simplifying the training objective of their models.

\begin{table*}[]
\centering
\begin{scriptsize}
\setlength{\tabcolsep}{5pt}
\begin{tabular}{@{}lcccccccccccccccc@{}}
\toprule
\multicolumn{1}{l|}{\multirow{2}{*}{\textbf{Method}}} & \multicolumn{4}{c|}{\textbf{Bedroom}}                                                  & \multicolumn{4}{c|}{\textbf{Dining room}}                                               & \multicolumn{4}{c|}{\textbf{Living room}}                                              & \multicolumn{4}{c}{\textbf{Library}}                              \\
\multicolumn{1}{l|}{}                                 & \textbf{CKL}  & \textbf{FID}  & \textbf{CIoU} & \multicolumn{1}{c|}{\textbf{NIRate}} & \textbf{CKL}  & \textbf{FID}   & \textbf{CIoU} & \multicolumn{1}{c|}{\textbf{NIRate}} & \textbf{CKL}  & \textbf{FID}  & \textbf{CIoU} & \multicolumn{1}{c|}{\textbf{NIRate}} & \textbf{CKL}  & \textbf{FID}  & \textbf{CIoU} & \textbf{NIRate} \\ \midrule
\multicolumn{17}{l}{\textit{Original dataset}}                                                                                                                                                                                                                                                                                                                                                        \\
\multicolumn{1}{l|}{LayoutGPT}                        & 7.47          & 27.4          & 4.12          & \multicolumn{1}{c|}{58.1}              & ---             & ---              & ---             & \multicolumn{1}{c|}{---}                 & 2.25          & 55.7          & 3.11          & \multicolumn{1}{c|}{41.1}              & ---             & ---             & ---             & ---                 \\
\multicolumn{1}{l|}{ATISS}                            & 3.67          & 29.4          & 2.28          & \multicolumn{1}{c|}{71.5}              & 2.64          & 32.2           & 2.11          & \multicolumn{1}{c|}{43.4}              & 2.21          & 31.6          & 1.95          & \multicolumn{1}{c|}{49.8}              & 10.2          & 51.2          & 2.01          & 61.9              \\
\multicolumn{1}{l|}{DiffuScene}                       & \textbf{2.18} & 27.0          & 1.32          & \multicolumn{1}{c|}{79.1}              & \textbf{1.33} & 37.4           & 1.64          & \multicolumn{1}{c|}{62.5}              & \textbf{1.30} & 38.5          & 1.31          & \multicolumn{1}{c|}{62.3}              & ---             & ---             & ---             & ---                 \\
\multicolumn{1}{l|}{Ours}                             & 2.23          & \textbf{26.3} & \textbf{1.08} & \multicolumn{1}{c|}{\textbf{80.6}}     & 2.61          & \textbf{31.7}  & \textbf{1.57} & \multicolumn{1}{c|}{\textbf{63.9}}     & 2.05          & \textbf{30.1} & \textbf{1.05} & \multicolumn{1}{c|}{\textbf{77.5}}     & \textbf{9.24} & \textbf{47.0} & \textbf{1.40} & \textbf{75.1}     \\ \midrule
\multicolumn{17}{l}{\textit{\datasetname dataset}}                                                                                                                                                                                                                                                                                                                                     \\
\multicolumn{1}{l|}{LayoutGPT}                        & 6.46          & 28.4          & 4.12          & \multicolumn{1}{c|}{58.1}              & ---             & ---              & ---             & \multicolumn{1}{c|}{---}                 & 2.60          & 50.9          & 3.11          & \multicolumn{1}{c|}{41.1}              & ---             & ---             & ---             & ---                 \\
\multicolumn{1}{l|}{ATISS}                            & 3.04          & 27.0          & 1.15          & \multicolumn{1}{c|}{80.2}              & 2.58          & 31.98          & 1.90          & \multicolumn{1}{c|}{52.3}              & 1.77          & 30.0          & 1.55          & \multicolumn{1}{c|}{55.5}              & 9.31          & 51.4          & 1.16          & 77.1              \\
\multicolumn{1}{l|}{DiffuScene} & 2.08 & 28.0 & 1.32 & \multicolumn{1}{c|}{79.1} & --- & --- & --- & \multicolumn{1}{c|}{---} & \textbf{1.22} & 35.7 & 1.12 & \multicolumn{1}{c|}{65.8} & --- & --- & --- & --- \\
\multicolumn{1}{l|}{Ours}                             & \textbf{1.68} & \textbf{24.7} & \textbf{0.77} & \multicolumn{1}{c|}{\textbf{90.6}}     & \textbf{1.75} & \textbf{31.36} & \textbf{0.92} & \multicolumn{1}{c|}{\textbf{81.4}}     & 1.59 & \textbf{28.4} & \textbf{0.62} & \multicolumn{1}{c|}{\textbf{89.0}}     & \textbf{8.28} & \textbf{43.0} & \textbf{0.79} & \textbf{87.1}     \\ \bottomrule
\end{tabular}
\end{scriptsize}
\caption{Quantitative comparison of various layout generation methods across multiple evaluation metrics, highlighting the performance advantages of our proposed approach. It should be noted that LayoutGPT is a training-free method using GPT-4 and only generates \textit{bedroom} and \textit{living room} scenes, so the calculation of CIoU and NIRate is not applicable to the dataset.}
\label{tab:main results}
\vspace{-1.0em}
\end{table*}

\subsection{Applications}

\begin{figure}
    \centering
    \includegraphics[width=1.0\linewidth]{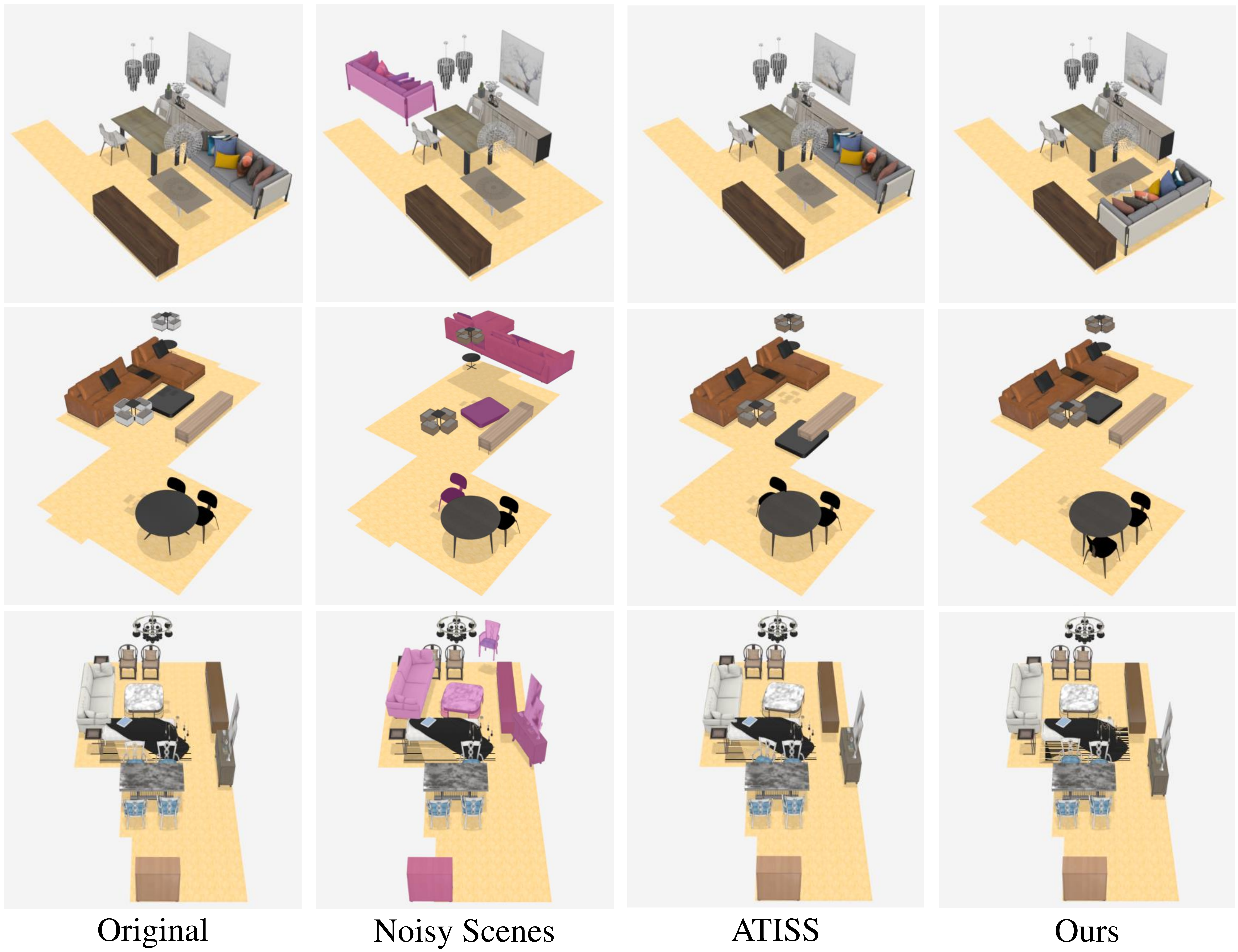}
\vspace{-0.5em}
    \caption{Comparison of scene re-arrangement across different methods. Magenta color is used to indicate disrupted objects.}
    \label{fig:scene_rearrange}
\vspace{-0.5em}
\end{figure}

\paragraph{Scene completion and scene re-arrangement} We evaluate our method on two tasks: scene completion and scene re-arrangement. For scene completion, we remove most objects from the scene, retain only a few key objects as ground truth, and predict the remaining tokens auto-regressively until completion. Our approach outperforms ATISS in coherence and effectively avoids object intersections, as visualized in the supplementary material. For scene re-arrangement, our method corrects failure cases by calculating each object's probability in the scene, selecting lower-probability objects, and resampling them. As demonstrated in \Cref{fig:scene_rearrange}, our approach successfully adjusts the positions of one, two, or multiple objects, achieving superior results compared to ATISS.

\subsection{Ablation study}
We conducted ablation studies to validate our design. First, we compared cuboid and bounding box-based intersection avoidance for dataset refinement. Next, we tested cuboid sequence modeling by training on both bounding box and cuboid sequences. 
Lastly, we ablated rejection sampling to show its effectiveness.

\begin{figure}[t]
\centering
\includegraphics[width=0.85\linewidth]{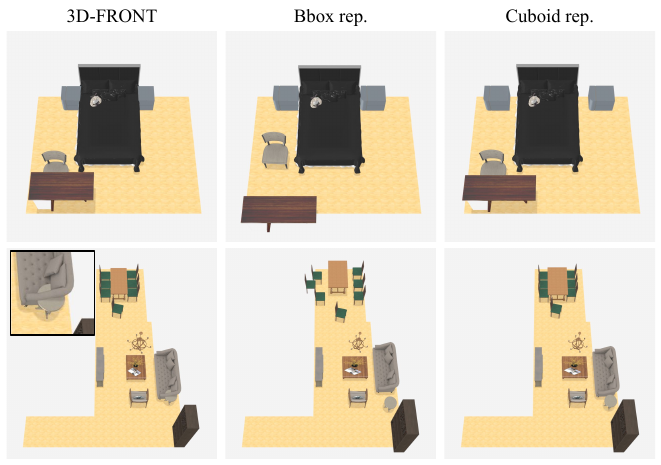}
\caption{Visual comparison of object arrangement results using different representations for dataset refinement.}
\label{fig:abl_datrefine}
\vspace{-1.0em}
\end{figure}

\vspace{0.5em}\noindent \textbf{Dataset refinement.}
To validate the superiority of using cuboid representation for dataset refinement, we also applied the intersection avoidance steps using the bounding box, following the same procedure in \Cref{{sec:dataset}}. \Cref{fig:abl_datrefine} shows the corresponding visual results, where it can be seen that while using bounding boxes does reduce intersections between objects, it causes previously non-intersecting arrangements, such as table-chair groups, to be unnecessarily spaced apart. In \Cref{tab:ablation refine}, we calculate the FID for both datasets refined with the cuboid and bounding box representation with the original 3D-FRONT dataset, as well as the NIRate of the two datasets. It can be seen that using cuboid representation achieves a higher NIRate while maintaining a smaller impact on the original data distribution.

\begin{table}[h]
\centering
\begin{scriptsize}
\setlength{\tabcolsep}{6pt}
\begin{tabular}{@{}l|cc|cc@{}}
\toprule
\multirow{2}{*}{\textbf{Scenes}} & \multicolumn{2}{c|}{\textbf{Bedroom}}         & \multicolumn{2}{c}{\textbf{Living room}}      \\
                                 & \textbf{Cuboid rep.} & \textbf{BBox rep.} & \textbf{Cuboid rep.} & \textbf{BBox rep.} \\ \midrule
FID                              & 4.43                   & 4.61                 & 16.3                   & 19.0                 \\
NIRate                         & 99.1                   & 96.3                 & 99.7                   & 99.2                 \\ \bottomrule
\end{tabular}
\end{scriptsize}
\caption{Ablation study on dataset intersection avoidance based on cuboid and bounding box representations. The cuboid representation achieves a higher NIRate with fewer perturbations (FID). }
\label{tab:ablation refine}
\vspace{-1.0em}
\end{table}

\noindent \textbf{Cuboid sequence modeling.}
We trained our sequence model using both bounding box and cuboid representations. Quantitative results in \Cref{tab:ablation cr} demonstrate the superiority of the cuboid representation, suggesting that extending sequence length within a reasonable range enhances the learning ability of sequence models.

\noindent \textbf{Rejection sampling.}
The lower part of \Cref{tab:ablation cr} shows the changes in model performance with each iteration of rejection sampling. With each iteration, both CIoU and NIRate steadily improve, demonstrating the effectiveness of our approach. However, FID experiences a slight decline in the final stages, as the distribution begins to diverge from the original dataset throughout training to a certain extent.

\begin{table}[t]
\centering
\begin{scriptsize}
\setlength{\tabcolsep}{6pt}
\begin{tabular}{@{}l|ccc|ccc@{}}
\toprule
\multirow{2}{*}{\textbf{Setting}} & \multicolumn{3}{c|}{\textbf{Bedroom}}            & \multicolumn{3}{c}{\textbf{Living room}}         \\
                                  & \textbf{FID} & \textbf{CIoU} & \textbf{NIRate} & \textbf{FID} & \textbf{CIoU} & \textbf{NIRate} \\ \midrule
BBox rep.                         & 28.6         & 1.93          & 71.8              & 38.9         & 2.11          & 49.2              \\
Cuboid rep.                       & 26.7         & 1.60          & 76.8              & 30.3         & 1.99          & 55.9              \\ 
\ \rule[0em]{0.4pt}{1em}$-$Reject. iter1                     & 25.7         & 1.37          & 80.3              & 29.0         & 1.84          & 72.5              \\
\ \rule[0em]{0.4pt}{1em}$-$Reject. iter2                     & 24.2         & 0.85          & 87.0              & 28.1         & 0.98          & 85.0              \\
\ \rule[0.4em]{0.4pt}{0.7em}$-$Reject. iter3                     & 24.7         & 0.77          & 90.6              & 28.4         & 0.62          & 89.0              \\ \bottomrule
\end{tabular}
\end{scriptsize}
\caption{Ablation study comparing bounding box \textit{vs.} cuboid representations (top) and rejection sampling iterations (bottom).}
\label{tab:ablation cr}
\vspace{-1.0em}
\end{table}

\section{Conclusion, limitations and future work}

In this work, we introduce \methodname, a novel framework for synthesizing indoor scenes by assembling cuboids. Unlike previous approaches, we tackle the challenge of object intersections in indoor scene synthesis by proposing a new cuboid representation. By integrating cuboid primitives with rejection sampling during training, our model significantly enhances the quality of scene generation, ensuring more coherent spatial arrangements and reducing unwanted intersections among objects. Despite these advances, certain limitations remain. In complex scenes with limited floor space, some objects may still shift off the floor plane, underscoring the need for more robust spatial awareness under constrained conditions. 

Looking ahead, future research direction could focus on exploring denoising diffusion models for refined cuboid generation, benefiting from their strong generation capabilities, as well as applying reinforcement learning to optimize the model, using intersection volumes as reward signals.

\section*{Acknowledgments}

This work was supported in part by the Natural Science Foundation of China under Grant  62372423, 62121002, U2336206, supported in part by the Anhui Province Key Laboratory of Digital Security and supported by GRF grant from the Research Grants Council (RGC) of the Hong Kong Special Administrative Region, China [Project No.  CityU 11208123], and was partly supported by NSERC Discovery, CFI-JELF, NSERC Alliance, Alberta Innovates and PrairiesCan grants. 
Additionally, we thank Jiyan He and Qi Sun for providing constructive suggestions.

{
    \small
    \bibliographystyle{ieeenat_fullname}
    \bibliography{main}

\begin{thebibliography}{56}
\providecommand{\natexlab}[1]{#1}
\providecommand{\url}[1]{\texttt{#1}}
\expandafter\ifx\csname urlstyle\endcsname\relax
  \providecommand{\doi}[1]{doi: #1}\else
  \providecommand{\doi}{doi: \begingroup \urlstyle{rm}\Url}\fi

\bibitem[Chang et~al.(2017)Chang, Dai, Funkhouser, Halber, Niebner, Savva, Song, Zeng, and Zhang]{chang2017matterport3d}
Angel Chang, Angela Dai, Thomas Funkhouser, Maciej Halber, Matthias Niebner, Manolis Savva, Shuran Song, Andy Zeng, and Yinda Zhang.
\newblock {Matterport3D}: Learning from {RGB-D} data in indoor environments.
\newblock In \emph{3DV}, 2017.

\bibitem[Cho et~al.(2025)Cho, Ivanovic, Cao, Schmerling, Wang, Weng, Li, You, Kr{\"a}henb{\"u}hl, Wang, et~al.]{cho2024language}
Jang~Hyun Cho, Boris Ivanovic, Yulong Cao, Edward Schmerling, Yue Wang, Xinshuo Weng, Boyi Li, Yurong You, Philipp Kr{\"a}henb{\"u}hl, Yan Wang, et~al.
\newblock Language-image models with {3D} understanding.
\newblock In \emph{ICLR}, 2025.

\bibitem[Deitke et~al.(2022)Deitke, VanderBilt, Herrasti, Weihs, Ehsani, Salvador, Han, Kolve, Kembhavi, and Mottaghi]{deitke2022️}
Matt Deitke, Eli VanderBilt, Alvaro Herrasti, Luca Weihs, Kiana Ehsani, Jordi Salvador, Winson Han, Eric Kolve, Aniruddha Kembhavi, and Roozbeh Mottaghi.
\newblock {ProcTHOR}: Large-scale embodied {AI} using procedural generation.
\newblock \emph{NeurIPS}, 2022.

\bibitem[Deng et~al.(2020)Deng, Genova, Yazdani, Bouaziz, Hinton, and Tagliasacchi]{deng2020cvxnet}
Boyang Deng, Kyle Genova, Soroosh Yazdani, Sofien Bouaziz, Geoffrey Hinton, and Andrea Tagliasacchi.
\newblock {CvxNet}: Learnable convex decomposition.
\newblock In \emph{CVPR}, 2020.

\bibitem[Dubey et~al.(2024)Dubey, Jauhri, Pandey, Kadian, Al-Dahle, Letman, Mathur, Schelten, Yang, Fan, et~al.]{dubey2024llama}
Abhimanyu Dubey, Abhinav Jauhri, Abhinav Pandey, Abhishek Kadian, Ahmad Al-Dahle, Aiesha Letman, Akhil Mathur, Alan Schelten, Amy Yang, Angela Fan, et~al.
\newblock The {Llama} 3 herd of models.
\newblock \emph{arXiv preprint arXiv:2407.21783}, 2024.

\bibitem[Feng et~al.(2023)Feng, Zhu, Fu, Jampani, Akula, He, Basu, Wang, and Wang]{feng2023layoutgpt}
Weixi Feng, Wanrong Zhu, Tsu-jui Fu, Varun Jampani, Arjun Akula, Xuehai He, Sugato Basu, Xin~Eric Wang, and William~Yang Wang.
\newblock Layout{GPT}: Compositional visual planning and generation with large language models.
\newblock In \emph{NeurIPS}, 2023.

\bibitem[Fisher et~al.(2011)Fisher, Savva, and Hanrahan]{fisher2011characterizing}
Matthew Fisher, Manolis Savva, and Pat Hanrahan.
\newblock Characterizing structural relationships in scenes using graph kernels.
\newblock In \emph{SIGGRAPH}, 2011.

\bibitem[Fisher et~al.(2012)Fisher, Ritchie, Savva, Funkhouser, and Hanrahan]{fisher2012example}
Matthew Fisher, Daniel Ritchie, Manolis Savva, Thomas Funkhouser, and Pat Hanrahan.
\newblock Example-based synthesis of {3D} object arrangements.
\newblock \emph{ACM TOG}, 2012.

\bibitem[Fisher et~al.(2015)Fisher, Savva, Li, Hanrahan, and Nie{\ss}ner]{fisher2015activity}
Matthew Fisher, Manolis Savva, Yangyan Li, Pat Hanrahan, and Matthias Nie{\ss}ner.
\newblock Activity-centric scene synthesis for functional {3D} scene modeling.
\newblock \emph{ACM TOG}, 2015.

\bibitem[Fu et~al.(2021{\natexlab{a}})Fu, Cai, Gao, Zhang, Wang, Li, Zeng, Sun, Jia, Zhao, et~al.]{fu20213front}
Huan Fu, Bowen Cai, Lin Gao, Ling-Xiao Zhang, Jiaming Wang, Cao Li, Qixun Zeng, Chengyue Sun, Rongfei Jia, Binqiang Zhao, et~al.
\newblock {3D-FRONT}: 3{D} furnished rooms with layouts and semantics.
\newblock In \emph{CVPR}, 2021{\natexlab{a}}.

\bibitem[Fu et~al.(2021{\natexlab{b}})Fu, Jia, Gao, Gong, Zhao, Maybank, and Tao]{fu20213d}
Huan Fu, Rongfei Jia, Lin Gao, Mingming Gong, Binqiang Zhao, Steve Maybank, and Dacheng Tao.
\newblock {3D-FUTURE}: 3{D} furniture shape with texture.
\newblock \emph{IJCV}, 2021{\natexlab{b}}.

\bibitem[Fu et~al.(2017)Fu, Chen, Wang, Wen, Zhou, and Fu]{fu2017adaptive}
Qiang Fu, Xiaowu Chen, Xiaotian Wang, Sijia Wen, Bin Zhou, and Hongbo Fu.
\newblock Adaptive synthesis of indoor scenes via activity-associated object relation graphs.
\newblock \emph{ACM TOG}, 2017.

\bibitem[Gao et~al.(2023)Gao, Sun, Mo, Lai, Guibas, and Yang]{gao2023scenehgn}
Lin Gao, Jia-Mu Sun, Kaichun Mo, Yu-Kun Lai, Leonidas~J Guibas, and Jie Yang.
\newblock {SceneHGN}: Hierarchical graph networks for {3D} indoor scene generation with fine-grained geometry.
\newblock \emph{IEEE TPAMI}, 2023.

\bibitem[Heusel et~al.(2017)Heusel, Ramsauer, Unterthiner, Nessler, and Hochreiter]{heusel2017gans}
Martin Heusel, Hubert Ramsauer, Thomas Unterthiner, Bernhard Nessler, and Sepp Hochreiter.
\newblock {GANs} trained by a two time-scale update rule converge to a local {Nash} equilibrium.
\newblock \emph{NeurIPS}, 2017.

\bibitem[Jiang et~al.(2012)Jiang, Lim, and Saxena]{jiang2012learning}
Yun Jiang, Marcus Lim, and Ashutosh Saxena.
\newblock Learning object arrangements in {3D} scenes using human context.
\newblock In \emph{ICML}, 2012.

\bibitem[Jones et~al.(2020)Jones, Barton, Xu, Wang, Jiang, Guerrero, Mitra, and Ritchie]{jones2020shapeassembly}
R~Kenny Jones, Theresa Barton, Xianghao Xu, Kai Wang, Ellen Jiang, Paul Guerrero, Niloy~J Mitra, and Daniel Ritchie.
\newblock {ShapeAssembly}: Learning to generate programs for {3D} shape structure synthesis.
\newblock \emph{ACM TOG}, 2020.

\bibitem[Khaki et~al.(2024)Khaki, Li, Ma, Yang, and Ramachandra]{khaki2024rs}
Saeed Khaki, JinJin Li, Lan Ma, Liu Yang, and Prathap Ramachandra.
\newblock {RS-DPO}: A hybrid rejection sampling and direct preference optimization method for alignment of large language models.
\newblock In \emph{NAACL}, 2024.

\bibitem[Kusner et~al.(2017)Kusner, Paige, and Hern{\'a}ndez-Lobato]{kusner2017grammar}
Matt~J Kusner, Brooks Paige, and Jos{\'e}~Miguel Hern{\'a}ndez-Lobato.
\newblock Grammar variational autoencoder.
\newblock In \emph{ICML}, 2017.

\bibitem[Leimer et~al.(2022)Leimer, Guerrero, Weiss, and Musialski]{leimer2022layoutenhancer}
Kurt Leimer, Paul Guerrero, Tomer Weiss, and Przemyslaw Musialski.
\newblock {LayoutEnhancer}: Generating good indoor layouts from imperfect data.
\newblock In \emph{SIGGRAPH Asia}, 2022.

\bibitem[Lewis et~al.(2020)Lewis, Liu, Goyal, Ghazvininejad, Mohamed, Levy, Stoyanov, and Zettlemoyer]{lewis2020bart}
Mike Lewis, Yinhan Liu, Naman Goyal, Marjan Ghazvininejad, Abdelrahman Mohamed, Omer Levy, Veselin Stoyanov, and Luke Zettlemoyer.
\newblock {BART}: Denoising sequence-to-sequence pre-training for natural language generation, translation, and comprehension.
\newblock In \emph{ACL}, 2020.

\bibitem[Li et~al.(2019)Li, Patil, Xu, Chaudhuri, Khan, Shamir, Tu, Chen, Cohen-Or, and Zhang]{li2019grains}
Manyi Li, Akshay~Gadi Patil, Kai Xu, Siddhartha Chaudhuri, Owais Khan, Ariel Shamir, Changhe Tu, Baoquan Chen, Daniel Cohen-Or, and Hao Zhang.
\newblock {GRAINS}: Generative recursive autoencoders for indoor scenes.
\newblock \emph{ACM TOG}, 2019.

\bibitem[Lin and Mu(2024)]{lininstructscene}
Chenguo Lin and Yadong Mu.
\newblock {InstructScene}: Instruction-driven {3D} indoor scene synthesis with semantic graph prior.
\newblock In \emph{ICLR}, 2024.

\bibitem[Liu et~al.(2024)Liu, Zhao, Joshi, Khalman, Saleh, Liu, and Liu]{liu2023statistical}
Tianqi Liu, Yao Zhao, Rishabh Joshi, Misha Khalman, Mohammad Saleh, Peter~J Liu, and Jialu Liu.
\newblock Statistical rejection sampling improves preference optimization.
\newblock In \emph{ICLR}, 2024.

\bibitem[Loshchilov and Hutter(2019)]{loshchilov2017decoupled}
Ilya Loshchilov and Frank Hutter.
\newblock Decoupled weight decay regularization.
\newblock In \emph{ICLR}, 2019.

\bibitem[Ma et~al.(2016)Ma, Li, Zou, Liao, Tong, and Zhang]{ma2016action}
Rui Ma, Honghua Li, Changqing Zou, Zicheng Liao, Xin Tong, and Hao Zhang.
\newblock Action-driven {3D} indoor scene evolution.
\newblock \emph{ACM TOG}, 2016.

\bibitem[Maillard et~al.(2024)Maillard, Sereyjol-Garros, Durand, and Ovsjanikov]{maillard2024debara}
L{\'e}opold Maillard, Nicolas Sereyjol-Garros, Tom Durand, and Maks Ovsjanikov.
\newblock {DeBaRA}: Denoising-based {3D} room arrangement generation.
\newblock \emph{NeurIPS}, 2024.

\bibitem[Merrell et~al.(2011)Merrell, Schkufza, Li, Agrawala, and Koltun]{merrell2011interactive}
Paul Merrell, Eric Schkufza, Zeyang Li, Maneesh Agrawala, and Vladlen Koltun.
\newblock Interactive furniture layout using interior design guidelines.
\newblock \emph{ACM TOG}, 2011.

\bibitem[Murphy(2012)]{murphy2012machine}
Kevin~P Murphy.
\newblock \emph{{Machine Learning}: {A} probabilistic perspective}.
\newblock MIT press, 2012.

\bibitem[Para et~al.(2023)Para, Guerrero, Mitra, and Wonka]{cofs}
Wamiq~Reyaz Para, Paul Guerrero, Niloy Mitra, and Peter Wonka.
\newblock {COFS}: Controllable furniture layout synthesis.
\newblock In \emph{SIGGRAPH}, 2023.

\bibitem[Paschalidou et~al.(2019)Paschalidou, Ulusoy, and Geiger]{paschalidou2019superquadrics}
Despoina Paschalidou, Ali~Osman Ulusoy, and Andreas Geiger.
\newblock {Superquadrics Revisited}: Learning {3D} shape parsing beyond cuboids.
\newblock In \emph{CVPR}, 2019.

\bibitem[Paschalidou et~al.(2021)Paschalidou, Kar, Shugrina, Kreis, Geiger, and Fidler]{paschalidou2021atiss}
Despoina Paschalidou, Amlan Kar, Maria Shugrina, Karsten Kreis, Andreas Geiger, and Sanja Fidler.
\newblock {ATISS}: Autoregressive transformers for indoor scene synthesis.
\newblock \emph{NeurIPS}, 2021.

\bibitem[Purkait et~al.(2020)Purkait, Zach, and Reid]{purkait2020sg}
Pulak Purkait, Christopher Zach, and Ian Reid.
\newblock {SG-VAE}: {Scene} {Grammar} {Variational} {Autoencoder} to generate new indoor scenes.
\newblock In \emph{ECCV}, 2020.

\bibitem[Qi et~al.(2018)Qi, Zhu, Huang, Jiang, and Zhu]{qi2018human}
Siyuan Qi, Yixin Zhu, Siyuan Huang, Chenfanfu Jiang, and Song-Chun Zhu.
\newblock Human-centric indoor scene synthesis using stochastic grammar.
\newblock In \emph{CVPR}, 2018.

\bibitem[Shazeer(2020)]{shazeer2020glu}
Noam Shazeer.
\newblock {GLU} variants improve transformer.
\newblock \emph{arXiv preprint arXiv:2002.05202}, 2020.

\bibitem[Shuai et~al.(2023)Shuai, Zhang, Yang, and Chen]{shuai2023dpf}
Qingyao Shuai, Chi Zhang, Kaizhi Yang, and Xuejin Chen.
\newblock {DPF-Net}: Combining explicit shape priors in deformable primitive field for unsupervised structural reconstruction of {3D} objects.
\newblock In \emph{ICCV}, 2023.

\bibitem[Su et~al.(2024)Su, Ahmed, Lu, Pan, Bo, and Liu]{su2024roformer}
Jianlin Su, Murtadha Ahmed, Yu Lu, Shengfeng Pan, Wen Bo, and Yunfeng Liu.
\newblock {RoFormer}: Enhanced transformer with rotary position embedding.
\newblock \emph{Neurocomputing}, 2024.

\bibitem[Sun et~al.(2024)Sun, Zhou, Zhou, Li, and Li]{sun2024forest2seq}
Qi Sun, Hang Zhou, Wengang Zhou, Li Li, and Houqiang Li.
\newblock {FOREST2SEQ}: Revitalizing order prior for sequential indoor scene synthesis.
\newblock In \emph{ECCV}, 2024.

\bibitem[Tang et~al.(2024)Tang, Nie, Markhasin, Dai, Thies, and Nie{\ss}ner]{tang2024diffuscene}
Jiapeng Tang, Yinyu Nie, Lev Markhasin, Angela Dai, Justus Thies, and Matthias Nie{\ss}ner.
\newblock {DiffuScene}: Denoising diffusion models for generative indoor scene synthesis.
\newblock In \emph{CVPR}, 2024.

\bibitem[Tulsiani et~al.(2017)Tulsiani, Su, Guibas, Efros, and Malik]{tulsiani2017learning}
Shubham Tulsiani, Hao Su, Leonidas~J Guibas, Alexei~A Efros, and Jitendra Malik.
\newblock Learning shape abstractions by assembling volumetric primitives.
\newblock In \emph{CVPR}, 2017.

\bibitem[Wang et~al.(2019)Wang, Lin, Weissmann, Savva, Chang, and Ritchie]{wang2019planit}
Kai Wang, Yu-An Lin, Ben Weissmann, Manolis Savva, Angel~X Chang, and Daniel Ritchie.
\newblock {PlanIT}: Planning and instantiating indoor scenes with relation graph and spatial prior networks.
\newblock \emph{ACM TOG}, 2019.

\bibitem[Wang et~al.(2021)Wang, Yeshwanth, and Nie{\ss}ner]{wang2021sceneformer}
Xinpeng Wang, Chandan Yeshwanth, and Matthias Nie{\ss}ner.
\newblock {SceneFormer}: Indoor scene generation with transformers.
\newblock In \emph{3DV}, 2021.

\bibitem[Wei et~al.(2023)Wei, Ding, Park, Sajnani, Poulenard, Sridhar, and Guibas]{wei2023lego}
Qiuhong~Anna Wei, Sijie Ding, Jeong~Joon Park, Rahul Sajnani, Adrien Poulenard, Srinath Sridhar, and Leonidas Guibas.
\newblock {LEGO-Net}: Learning regular rearrangements of objects in rooms.
\newblock In \emph{CVPR}, 2023.

\bibitem[Weiss et~al.(2018)Weiss, Litteneker, Duncan, Nakada, Jiang, Yu, and Terzopoulos]{weiss2018fast}
Tomer Weiss, Alan Litteneker, Noah Duncan, Masaki Nakada, Chenfanfu Jiang, Lap-Fai Yu, and Demetri Terzopoulos.
\newblock Fast and scalable position-based layout synthesis.
\newblock \emph{IEEE TVCG}, 2018.

\bibitem[Xu et~al.(2002)Xu, Stewart, and Fiume]{xu2002constraint}
Ken Xu, James Stewart, and Eugene Fiume.
\newblock Constraint-based automatic placement for scene composition.
\newblock In \emph{Graphics Interface}, 2002.

\bibitem[Yang and Chen(2021)]{yang2021unsupervised}
Kaizhi Yang and Xuejin Chen.
\newblock Unsupervised learning for cuboid shape abstraction via joint segmentation from point clouds.
\newblock \emph{ACM TOG}, 2021.

\bibitem[Yang et~al.(2024)Yang, Jia, Zhi, and Huang]{yang2024physcene}
Yandan Yang, Baoxiong Jia, Peiyuan Zhi, and Siyuan Huang.
\newblock {PHYSCENE}: Physically interactable {3D} scene synthesis for embodied {AI}.
\newblock In \emph{CVPR}, 2024.

\bibitem[Yin et~al.(2021)Yin, Zhou, and Krahenbuhl]{yin2021center}
Tianwei Yin, Xingyi Zhou, and Philipp Krahenbuhl.
\newblock Center-based {3D} object detection and tracking.
\newblock In \emph{CVPR}, 2021.

\bibitem[Yu et~al.(2011)Yu, Yeung, Tang, Terzopoulos, Chan, and Osher]{yu2011make}
Lap-Fai Yu, Sai~Kit Yeung, Chi~Keung Tang, Demetri Terzopoulos, Tony~F Chan, and Stanley~J Osher.
\newblock {Make It Home}: automatic optimization of furniture arrangement.
\newblock \emph{ACM TOG}, 2011.

\bibitem[Yu et~al.(2015)Yu, Yeung, and Terzopoulos]{yu2015clutterpalette}
Lap-Fai Yu, Sai-Kit Yeung, and Demetri Terzopoulos.
\newblock The {Clutterpalette}: An interactive tool for detailing indoor scenes.
\newblock \emph{IEEE TVCG}, 2015.

\bibitem[Yuan et~al.(2023)Yuan, Yuan, Li, Dong, Lu, Tan, Zhou, and Zhou]{yuan2023scaling}
Zheng Yuan, Hongyi Yuan, Chengpeng Li, Guanting Dong, Keming Lu, Chuanqi Tan, Chang Zhou, and Jingren Zhou.
\newblock Scaling relationship on learning mathematical reasoning with large language models.
\newblock \emph{arXiv preprint arXiv:2308.01825}, 2023.

\bibitem[Zhai et~al.(2024)Zhai, {\"O}rnek, Wu, Di, Tombari, Navab, and Busam]{zhai2024commonscenes}
Guangyao Zhai, Evin~P{\i}nar {\"O}rnek, Shun-Cheng Wu, Yan Di, Federico Tombari, Nassir Navab, and Benjamin Busam.
\newblock {CommonScenes}: Generating commonsense {3D} indoor scenes with scene graphs.
\newblock \emph{NeurIPS}, 2024.

\bibitem[Zhai et~al.(2025)Zhai, {\"O}rnek, Chen, Liao, Di, Navab, Tombari, and Busam]{zhai2025echoscene}
Guangyao Zhai, Evin~P{\i}nar {\"O}rnek, Dave~Zhenyu Chen, Ruotong Liao, Yan Di, Nassir Navab, Federico Tombari, and Benjamin Busam.
\newblock {EchoScene}: Indoor scene generation via information echo over scene graph diffusion.
\newblock In \emph{ECCV}, 2025.

\bibitem[Zhang and Sennrich(2019)]{zhang2019root}
Biao Zhang and Rico Sennrich.
\newblock Root mean square layer normalization.
\newblock \emph{NeurIPS}, 2019.

\bibitem[Zhao et~al.(2021)Zhao, Lin, Jia, Gao, Thattai, Thomason, and Sukhatme]{zhao2021luminous}
Yizhou Zhao, Kaixiang Lin, Zhiwei Jia, Qiaozi Gao, Govind Thattai, Jesse Thomason, and Gaurav~S Sukhatme.
\newblock Luminous: Indoor scene generation for embodied ai challenges.
\newblock \emph{arXiv preprint arXiv:2111.05527}, 2021.

\bibitem[Zhou et~al.(2019{\natexlab{a}})Zhou, Fang, Song, Guan, Yin, Dai, and Yang]{zhou2019iou}
Dingfu Zhou, Jin Fang, Xibin Song, Chenye Guan, Junbo Yin, Yuchao Dai, and Ruigang Yang.
\newblock {IoU} loss for {2D/3D} object detection.
\newblock In \emph{3DV}, 2019{\natexlab{a}}.

\bibitem[Zhou et~al.(2019{\natexlab{b}})Zhou, While, and Kalogerakis]{zhou2019scenegraphnet}
Yang Zhou, Zachary While, and Evangelos Kalogerakis.
\newblock {SceneGraphNet}: Neural message passing for 3{D} indoor scene augmentation.
\newblock In \emph{CVPR}, 2019{\natexlab{b}}.

\end{thebibliography}
}

\clearpage
\setcounter{page}{1}
\maketitlesupplementary
\appendix

\noindent In this supplementary material, we include the following details: implementation specifics in~\Cref{supp:Implementations}, dataset-related statistics in~\Cref{supp:Dataset}, computation cost in~\Cref{supp:compute}, cuboid generation procedure evaluation in~\Cref{supp:cuboid gen}, some failure cases of \methodname in~\Cref{supp:Failure} and further visual results~\Cref{supp:More}.

\section{Implementations}\label{supp:Implementations}

\subsection{Model architecture}
The model architecture is based on LLaMA-3 with modifications to the model size and position encoding. Specifically, the model consists of 8 layers, with a hidden dimension of 512 and a parameter size of 27.4 million. It uses the SwiGLU activation function and a learned position encoding. Additionally, both the number of attention heads (\textit{Head n}) and key-value heads (\textit{KV head}) are set to 8.

\subsection{Cuboid merging algorithm}\label{supp:merging algorithm}

\paragraph{Voxel space segmentation.}
We describe the detailed algorithm for the initial simple segmentation on the target voxel $O$ here. 

We start by studying how to cover the binary voxel representation using at most $k$ rectangles while minimizing the total volume. Given a binary 3D array where 1 represents the target shape and 0 represents the background, we begin by identifying the minimum bounding box that encloses all target voxels.

We then iteratively refine this initial box through a dynamic programming approach. At each iteration, we identify the rectangle that, when subdivided, would eliminate the largest contiguous region of background voxels. 

The subdivision process depends on the position of the removed background region. When removing a background region from a corner, the remaining volume is partitioned into 2 new rectangles. When removed from an edge (but not a corner), the remaining volume is partitioned into 3 new rectangles. When removed from the interior (not touching any edge), the remaining volume is divided into 4 new rectangles, splitting along horizontal or vertical planes. This position-dependent subdivision optimally preserves the target geometry while eliminating empty spaces.

The algorithm considers both horizontal-first and vertical-first splitting strategies, selecting the one that better minimizes total volume. The iterative refinement continues until either the maximum allowed number of rectangles $k$ is reached or no further optimization is possible. This results in a compact representation that closely approximates the original voxel shape while maintaining a controlled number of segments. This initial segmentation serves as the foundation for subsequent refinement steps in our pipeline.

\paragraph{Dynamic threshold.}

To encourage the merging of smaller cuboids while penalizing the merging of larger ones, we introduce a dynamic threshold mechanism.
The dynamic threshold $\tau_{\text{dynamic}}$ is computed as a function of the merged cuboid’s volume, using the following formula:

\begin{equation}
\tau_{\text{dynamic}}(V_C) = \tau_{\text{min}} + (\tau_{\text{max}} - \tau_{\text{min}}) \cdot \exp\left(-\frac{V_C}{S}\right),  \label{eq:dynamic}
\end{equation}

where $\tau_{\text{min}}$ and $\tau_{\text{max}}$ are the minimum and maximum threshold values, $V_C$ is the volume of the merged cuboid $C$, and $S$ is a scaling parameter that controls the rate of exponential decay.

The merging algorithm proceeds iteratively. We attempt to merge each pair of cuboids using the predefined threshold (either static or dynamic). If merging is successful, the two cuboids are replaced by the merged cuboid, and the process continues until no further merges are possible.

We provide an algorithm to illustrate the cuboid merging process, which plays a key role in maintaining an efficient representation of 3D scenes.
The pseudocode for the algorithm is presented in~\cref{alg:simplified_merge}.

\begin{algorithm}
\caption{Cuboid Merging with Dynamic Threshold}
\label{alg:simplified_merge}
\begin{algorithmic}[1]
    \STATE \textbf{Input:} List of cuboids $C = \{C_1, C_2, \dots, C_n\}$, initial threshold $\tau_{\text{init}}$, dynamic scaling factor $S$
    \STATE \textbf{Output:} Merged cuboids list $\hat{C}$
    
    \STATE Initialize $\hat{C} \gets C$
    \STATE Set \textit{changed} to \texttt{True}
    
    \WHILE{\textit{changed}}
        \STATE Set \textit{changed} to \texttt{False}
        
        \STATE Select pairs of adjacent cuboids from $\hat{C}$
        \FOR{each selected pair $(C_i, C_j)$}
            \STATE Calculate the bounding cuboid $C_{\text{bounding}}$
            \STATE Compute volumes $V_{C_i}$, $V_{C_j}$, and $V_{C_{\text{bounding}}}$
            \STATE Calculate dynamic threshold $\tau_{\text{dynamic}}(V_{\text{C}})$
            \IF{$\frac{V_{\text{C}}}{V_{\text{A}} + V_{\text{B}}} < \tau_{\text{dynamic}}$}
                \STATE Merge $C_i$ and $C_j$ into $C_{\text{bounding}}$
                \STATE Remove $C_j$ from $\hat{C}$
                \STATE Set \textit{changed} to \texttt{True}
                \STATE \textbf{break} from the loop once a merge occurs
            \ENDIF
        \ENDFOR
    \ENDWHILE
    
    \STATE \textbf{return} $\hat{C}$
\end{algorithmic}
\end{algorithm}

\section{Dataset}\label{supp:Dataset}

\subsection{Failure cases in 3D-FRONT}
\begin{figure}[h]
\centering
\includegraphics[width=1.0\linewidth]{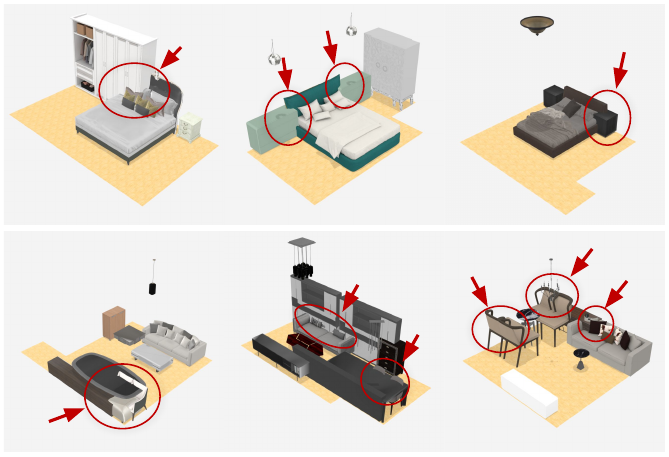}
\caption{Examples of object intersections in the 3D-FRONT dataset, with collided objects highlighted in red circles.}
\label{fig:3dfront failure}
\end{figure}

As described in the main text, the 3D-FRONT dataset contains numerous object intersections. As shown in~\Cref{fig:3dfront failure}, the 3D-FRONT dataset contains numerous examples of object intersections. These intersections can occur between various objects, such as tables and chairs, beds and nightstands, as well as wardrobes.

\begin{figure*}[t!]
\centering
\includegraphics[width=1.0\linewidth]{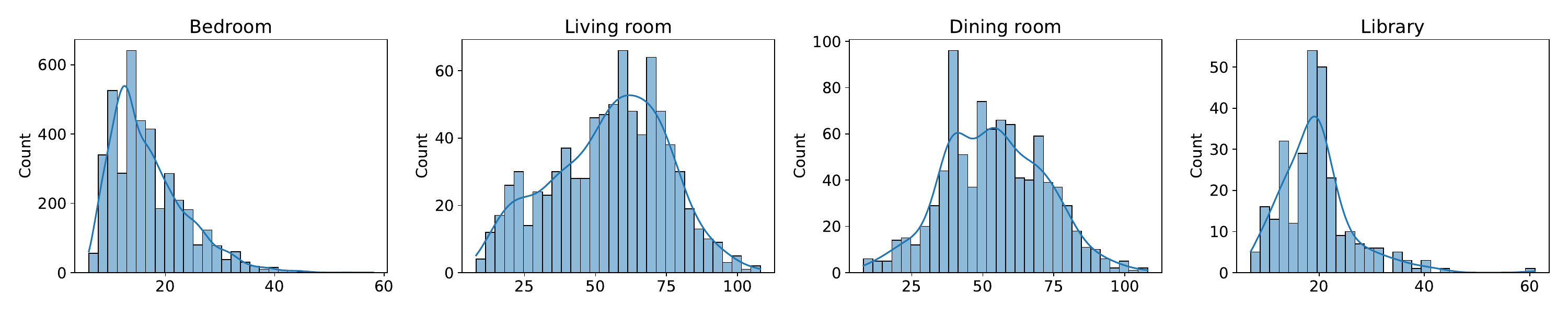}
\caption{Distribution of cuboid representation lengths in different room scenes. The histograms depict the frequency of cuboid lengths in bedroom, living room, dining room, and library scenes, showcasing the variation in cuboid sizes across these environments.}
\label{fig:cuboid scene}
\end{figure*}

\subsection{3D-FRONT and \datasetname comparison}
\begin{figure}[h]
\centering
\includegraphics[width=1.0\linewidth]{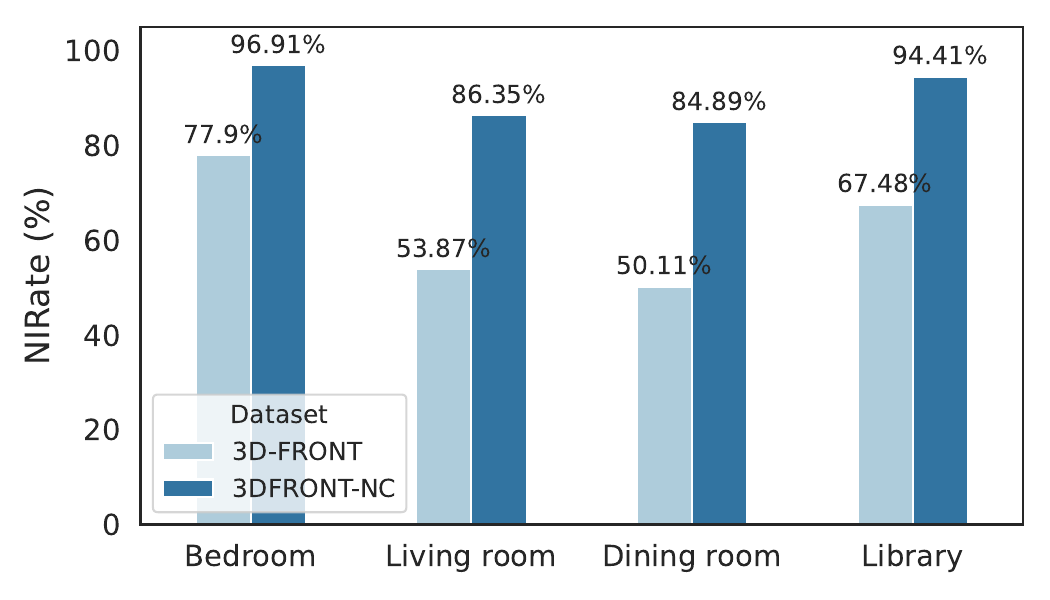}
\caption{Comparison of non-intersection rates (NIRate) between the 3D-FRONT and 3D-FRONT-NC datasets across different room types. The curated 3D-FRONT-NC dataset exhibits significantly higher NIRate, indicating fewer object intersections, particularly in bedrooms and libraries.}
\label{fig:room rate}
\end{figure}

We computed the non-intersection rate (NIRate) for both the 3D-FRONT and 3D-FRONT-NC datasets. \Cref{fig:room rate} illustrates our findings, showing that our curated dataset, 3D-FRONT-NC, has significantly fewer object intersections.

\subsection{Distributions of \datasetname}

\begin{figure}[h]
\centering
\includegraphics[width=1.0\linewidth]{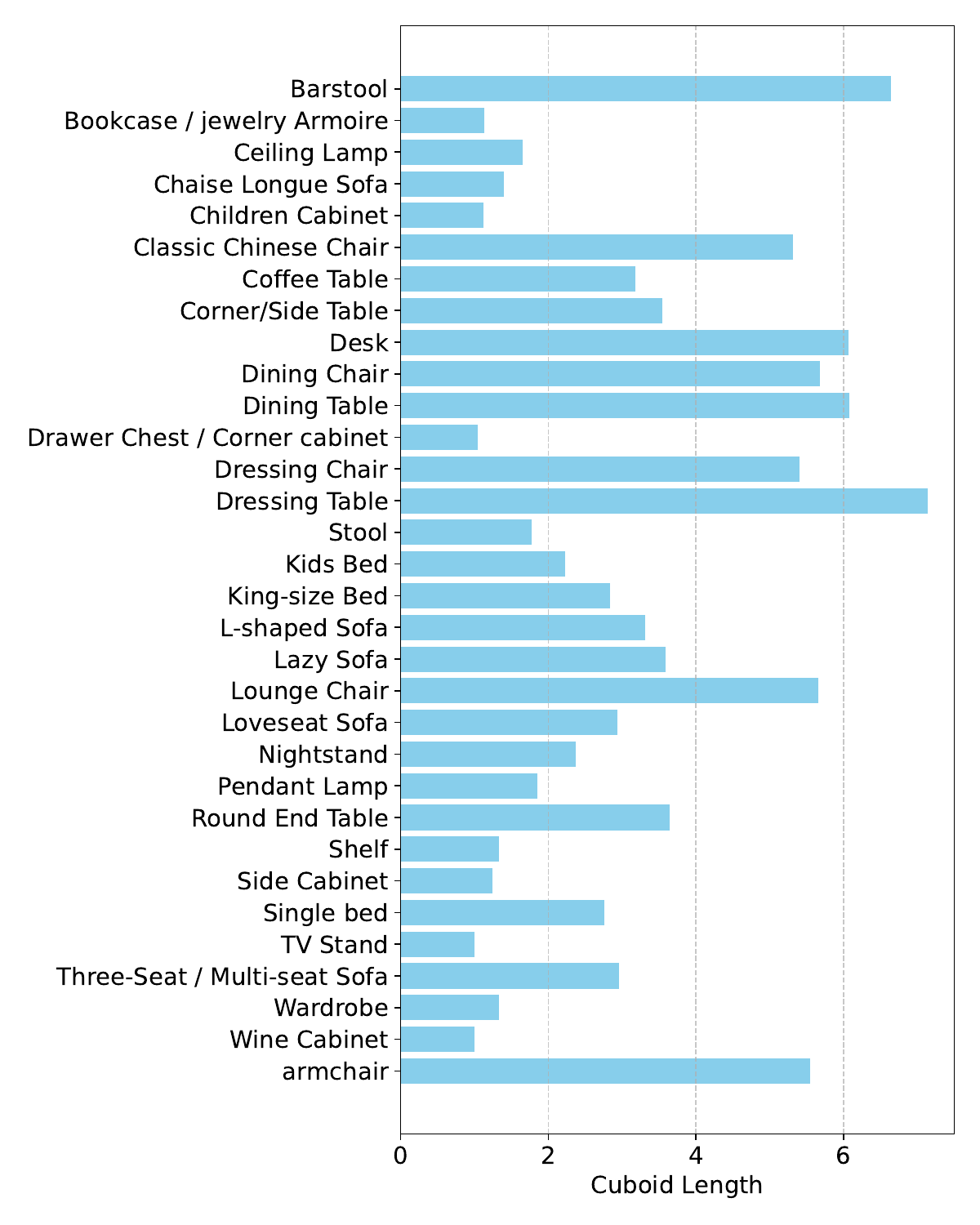}
\caption{Average cuboid length for different furniture categories. Note that ``\textit{footstool / sofa stool / bed end stool / stool}" is simplified as ``\textit{stool}", ``\textit{lounge chair / cafe chair / office chair}" as ``\textit{lounge chair}", and ``\textit{sideboard / side cabinet / console table}" as ``side \textit{cabinet}".}
\vspace{-1em}
\label{fig:cuboid len}
\end{figure}

\Cref{fig:cuboid len} illustrates the average cuboid lengths across various furniture categories. The category related to desks exhibits the longest average cuboid length, highlighting the inherent complexity of desks.

We also analyzed the cuboid representation lengths in bedroom, living room, dining room, and library scenes and plotted their distributions. \Cref{fig:cuboid scene} illustrates the distribution of cuboid lengths in each scene. 
It can be observed that the living room and dining room have longer cuboid lengths, likely due to the higher number of objects in these scenes.

\section{Computation cost}\label{supp:compute}

We compare the model size, inference time per forward pass, as well as the GPU time and GPU memory consumption during training of our method with others in \Cref{tab:comp cost}.

Additionally, We measured the average object retrieval time over 1,000 sampled scenes and compared it with ATISS and DiffuScene. The average retrieval time per object for ATISS, DiffuScene, and our method is 0.57s, 0.45s, and 0.70s, respectively, showing that our approach adds only a small overhead.

\begin{table}[h]
\centering
\begin{scriptsize}
\setlength{\tabcolsep}{3pt}
\begin{tabular}{@{}l|c|c|cc|c@{}}
\toprule
Methods      & Params (M) & Forward time & GPU hours & GPU Mem. & Retr. (s/obj) \\ \midrule
ATISS        & 8.4   & 36.68 ms      & 190 h       & 1.99 G    & 0.57         \\
DiffuScene   & 77.6  & 41.12 ms      & 185 h       & 3.69 G    & 0.57         \\
Ours         & 27.4 & 29.63 ms      & 55 h        & 11.8 G    & 0.70         \\
Ours w/ rej. & 27.4 & 29.63 ms      & 175 h       & 11.8 G    & 0.70         \\ \bottomrule
\end{tabular}
\end{scriptsize}
\caption{Comparison of computation cost for different methods.}
\label{tab:comp cost}
\end{table}

\section{Cuboid generation procedure evaluation}\label{supp:cuboid gen}
During the initial experiments, we explored some shape abstraction methods, but we found that some methods~\cite{deng2020cvxnet,paschalidou2019superquadrics} do not meet our goal of modeling with cuboids, and some methods are trained on specific ShapeNet categories and lack sufficient generalization to 3D-FRONT. Therefore, we designed our own cuboid shape abstraction method that works well on 3D-FRONT.

\begin{figure*}[t]
\centering
\includegraphics[width=0.8\linewidth]{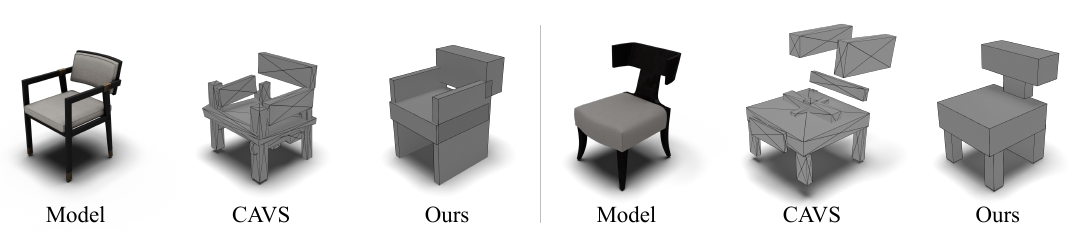}
\caption{CAVS performs reasonably well on relatively in-distribution models (left), but 3D-FRONT contains many OOD models (right), where CAVS struggles.}
\label{fig:cavs}
\end{figure*}

We compare our method with a relatively recent work~\cite{yang2021unsupervised} on 3D-FRONT. For convenience, we refer to it as CAVS. We use the authors' checkpoint trained on ShapeNet Chair and test it on 3D-FRONT Dining Chair. Our results, shown in \cref{tab:cavs} and \cref{fig:cavs}, demonstrate superior performance.

\begin{table}[h]
\centering
\begin{scriptsize}
\setlength{\tabcolsep}{4pt}
\begin{tabular}{l|ccc}
\toprule
Method & ChamferDistance-$L_1$ $\downarrow$ & ChamferDistance-\textit{fscore} $\uparrow$ & 3D-IoU $\uparrow$ \\ 
\midrule
CAVS. \cite{yang2021unsupervised} & 0.013 & 0.331 & 0.246 \\
Ours & 0.008 & 0.412 & 0.368 \\ 
\bottomrule
\end{tabular}
\end{scriptsize}
\caption{Comparison between our shape abstraction method with CAVS. Our shape abstraction method does not rely on neural networks, yet still achieves significant improvements. }
\label{tab:cavs}
\end{table}

\section{Failure cases of \methodname}\label{supp:Failure}

\paragraph{Cuboid generation plausibility.}

\begin{figure*}[h]
\centering
\includegraphics[width=0.8\linewidth]{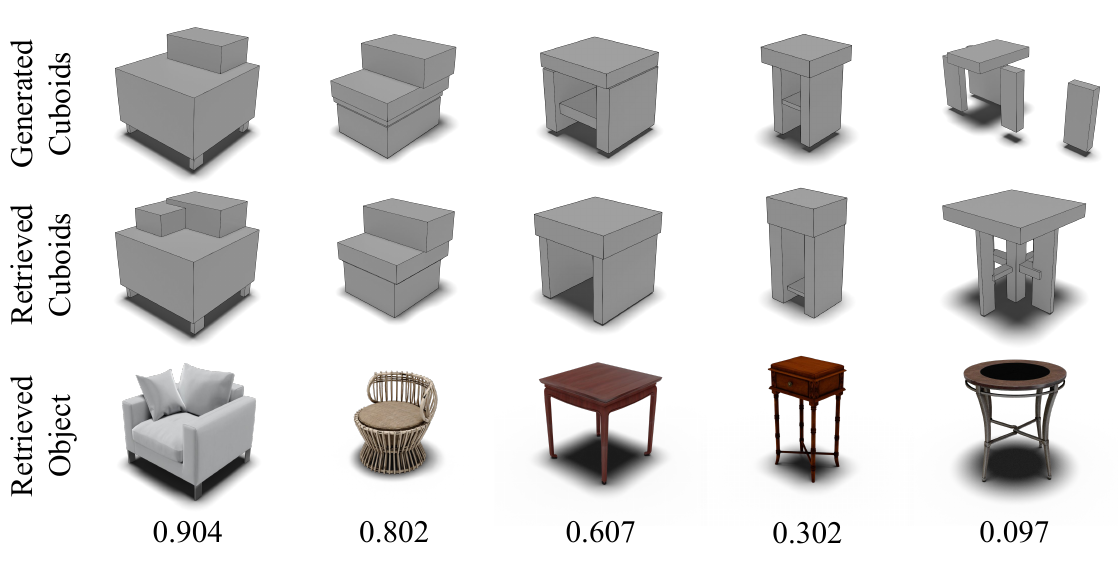}
\caption{Cuboid assemblies and retrieved objects under different max 3D IoU conditions.}
\label{fig:obj ret}
\end{figure*}

We perform object retrieval by computing 3D IoU on voxel grids. For a generated object's cuboid assembly, we compute its OBB box, align it to the origin, and select the candidate with the highest 3D IoU. \cref{fig:obj ret} highlights extreme anomaly cases with very low max 3D IoU. In most cases, our cuboid sequences resemble reasonable objects and failure cases are rare.

\paragraph{Cross boundary.}

\begin{figure*}[t]
\centering
\includegraphics[width=0.85\linewidth]{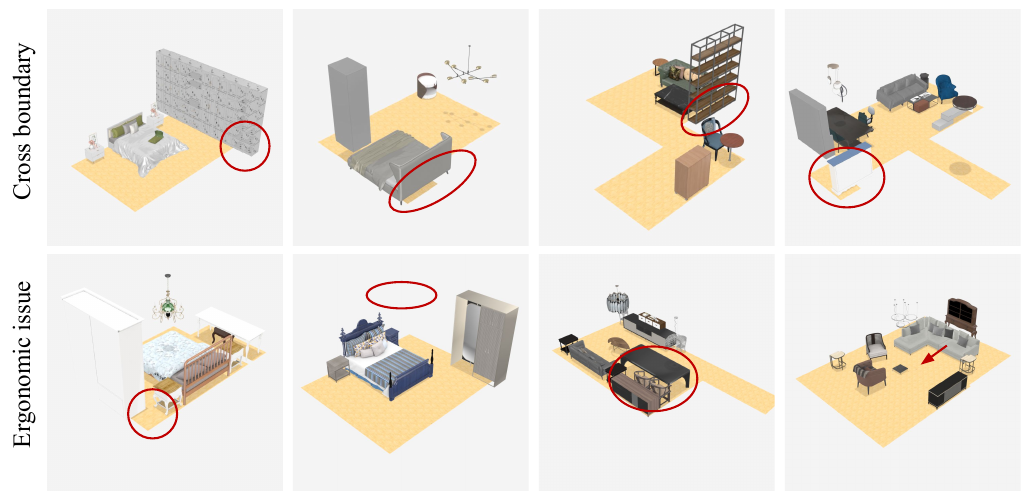}
\caption{Failure cases of our method, including boundary crossing (first row) and ergonomic issues (second row).}
\label{fig:failure case our}
\end{figure*}

\Cref{fig:failure case our} presents some failure cases of our method. Firstly, boundary constraints were not explicitly considered during training, leading to results where some objects exceed the floor boundaries (first row).

\paragraph{Ergonomic issue.}

Secondly, while we aim for the generated results to comply with \textit{ergonomic} principles, our model occasionally produces results that do not meet these standards. For instance, in the first column of the second row, there are areas on the floor that are inaccessible to people and disconnected from other floor areas. In the second column of the second row, there is no lighting. In the third column of the second row, a chair is placed in an inaccessible location. In the fourth column of the second row, the sofa's main orientation does not face the TV cabinet.

\section{More visual results}\label{supp:More}

\begin{figure*}
\centering
\includegraphics[width=0.85\linewidth]{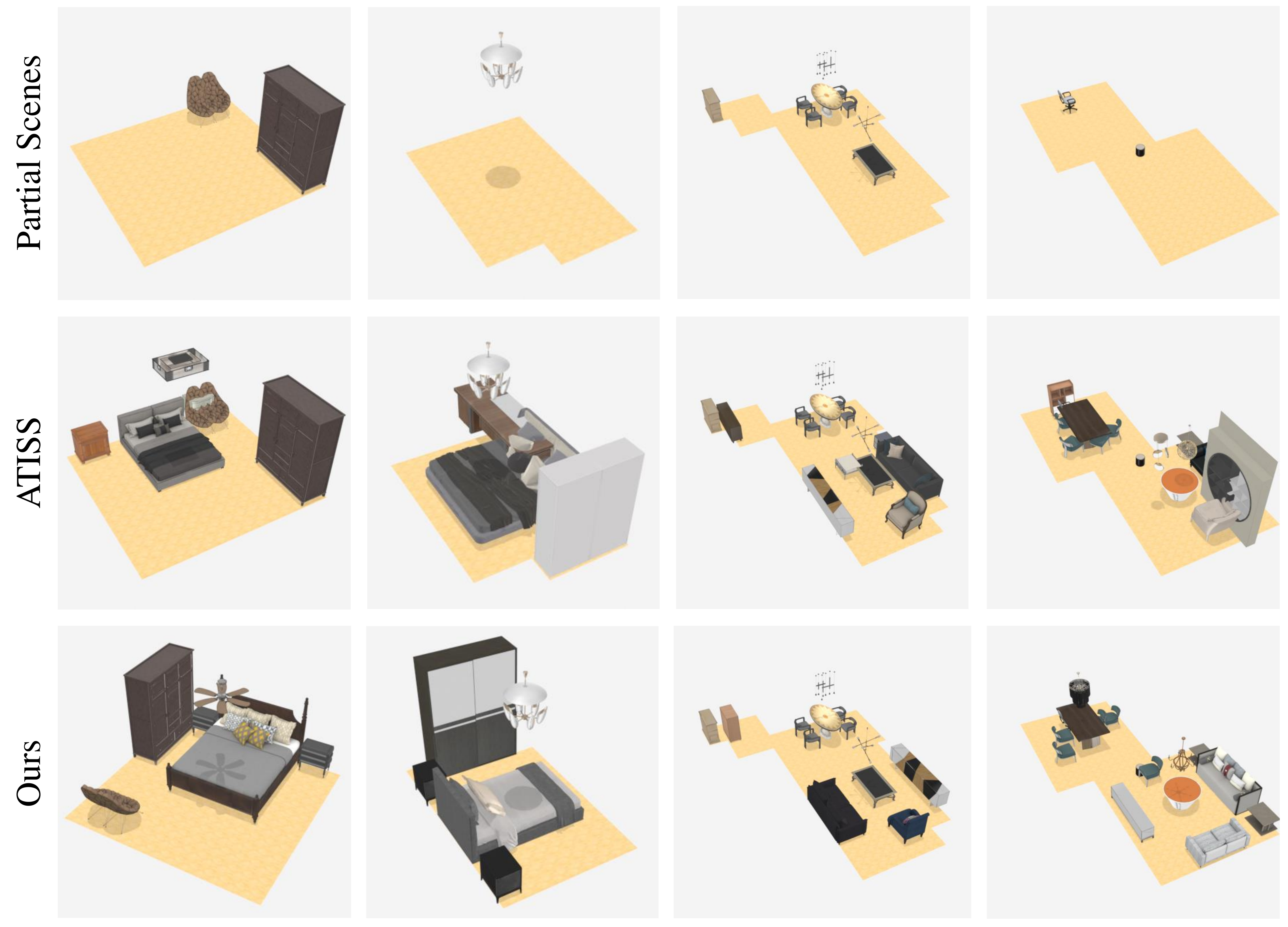}
\caption{Comparison of scene completion results between ATISS and our method. We present results for the \textit{bedroom} scene (left two columns) and the \textit{living room} scene (right two columns).}
\label{fig:scene_comple}
\end{figure*}

\begin{figure*}[t]
\centering
\includegraphics[width=0.95\linewidth]{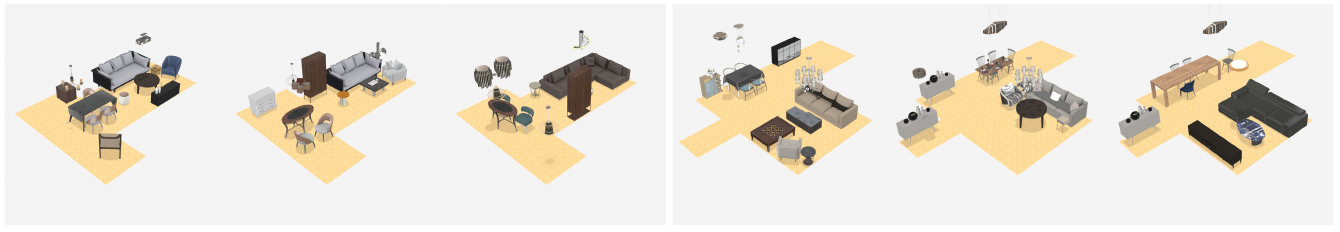}
\caption{Sampling results for 2 different floor plans in the living room scene.}
\label{fig:output diversity}
\end{figure*}

\begin{figure*}
\centering
\includegraphics[width=0.95\linewidth]{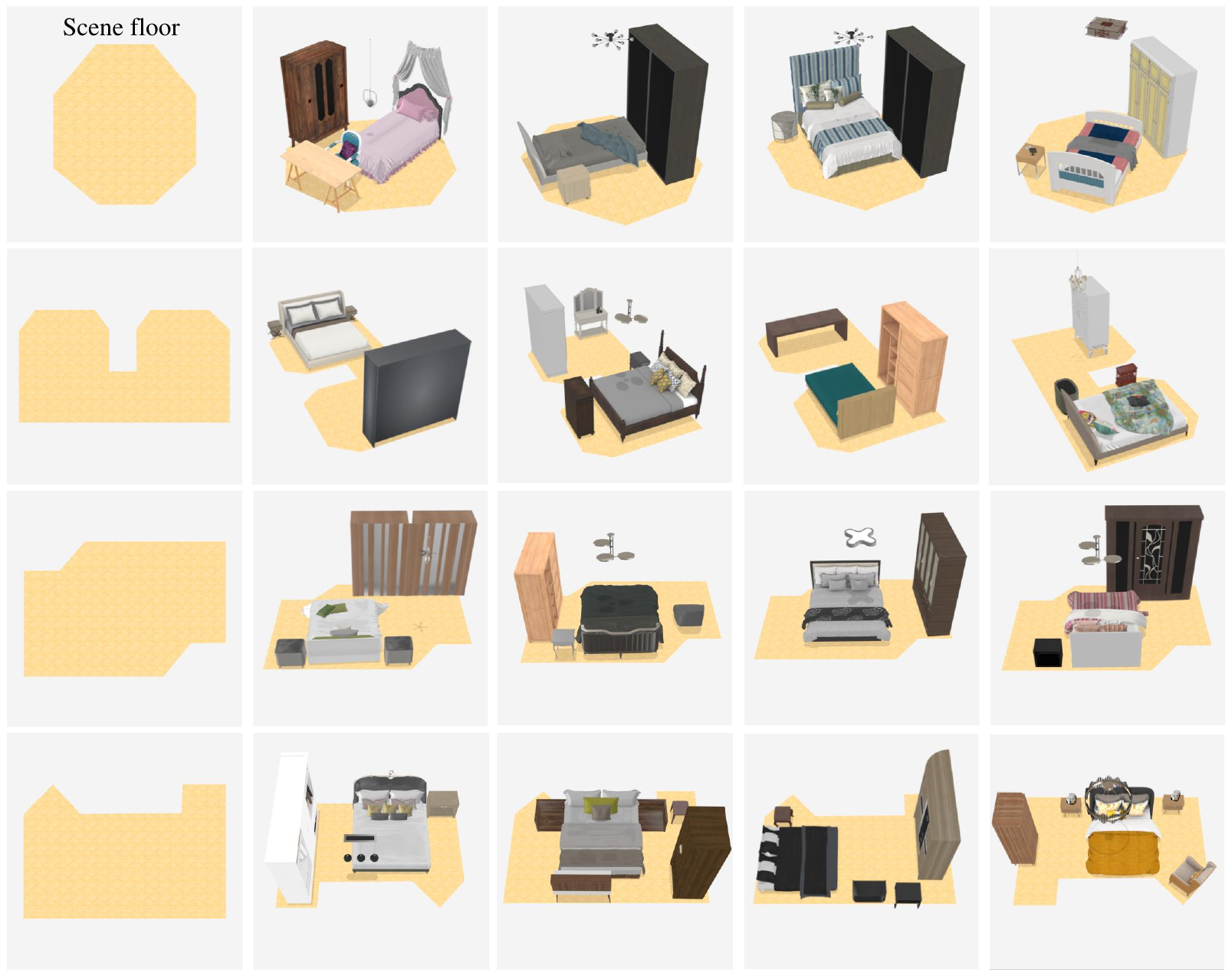}
\caption{Generalization beyond training data. We present four synthesized bedrooms generated based on four customized room layouts using our model.}
\label{fig:more_floor}
\end{figure*}

\begin{figure*}[t]
\centering
\includegraphics[width=0.95\linewidth]{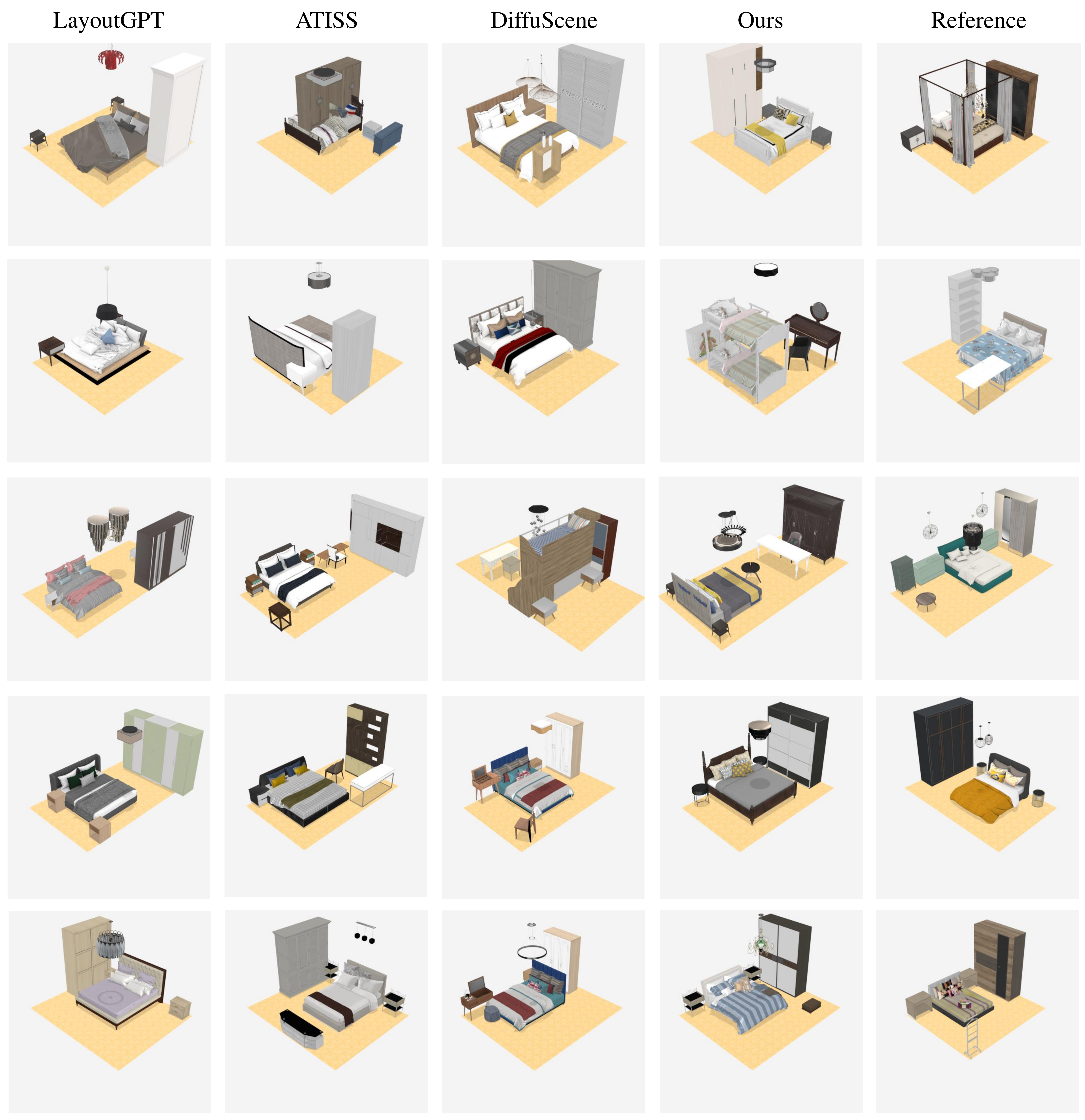}
\caption{Additional results of \textit{bedroom} scene synthesis. We compare our method with the state-of-the-art methods, where our results present better plausibility with fewer object collision issues.
}
\label{fig:add_cmp1}
\end{figure*}

\begin{figure*}[t]
\centering
\includegraphics[width=0.95\linewidth]{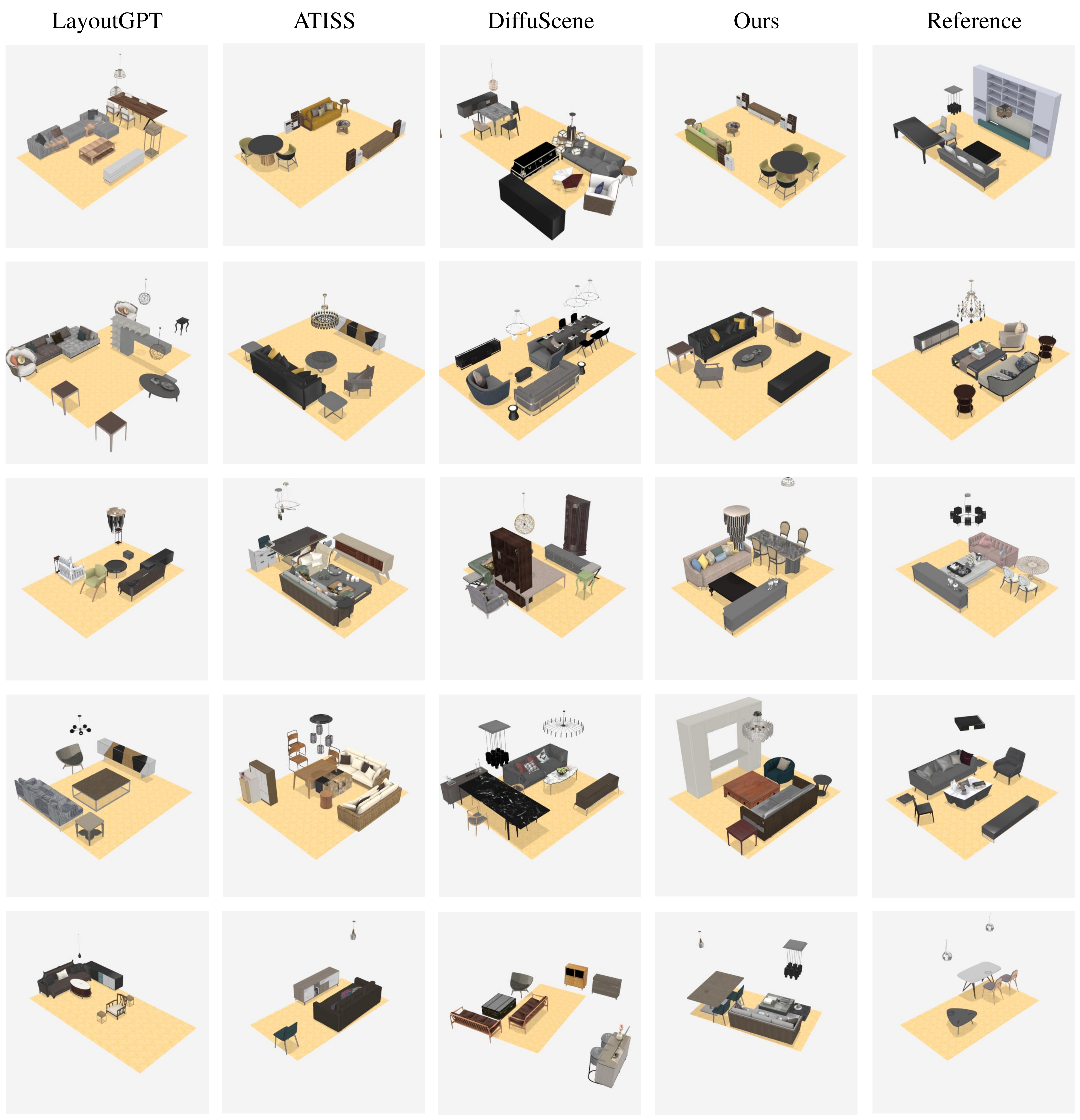}
\caption{Additional results of \textit{living room / dining room} scene synthesis show that compared to the state-of-the-art methods, our approach has fewer object collision issues, especially in the pairing of tables and chairs.
}
\label{fig:add_cmp2}
\end{figure*}

\begin{figure*}[t]
\centering
\includegraphics[width=0.95\linewidth]{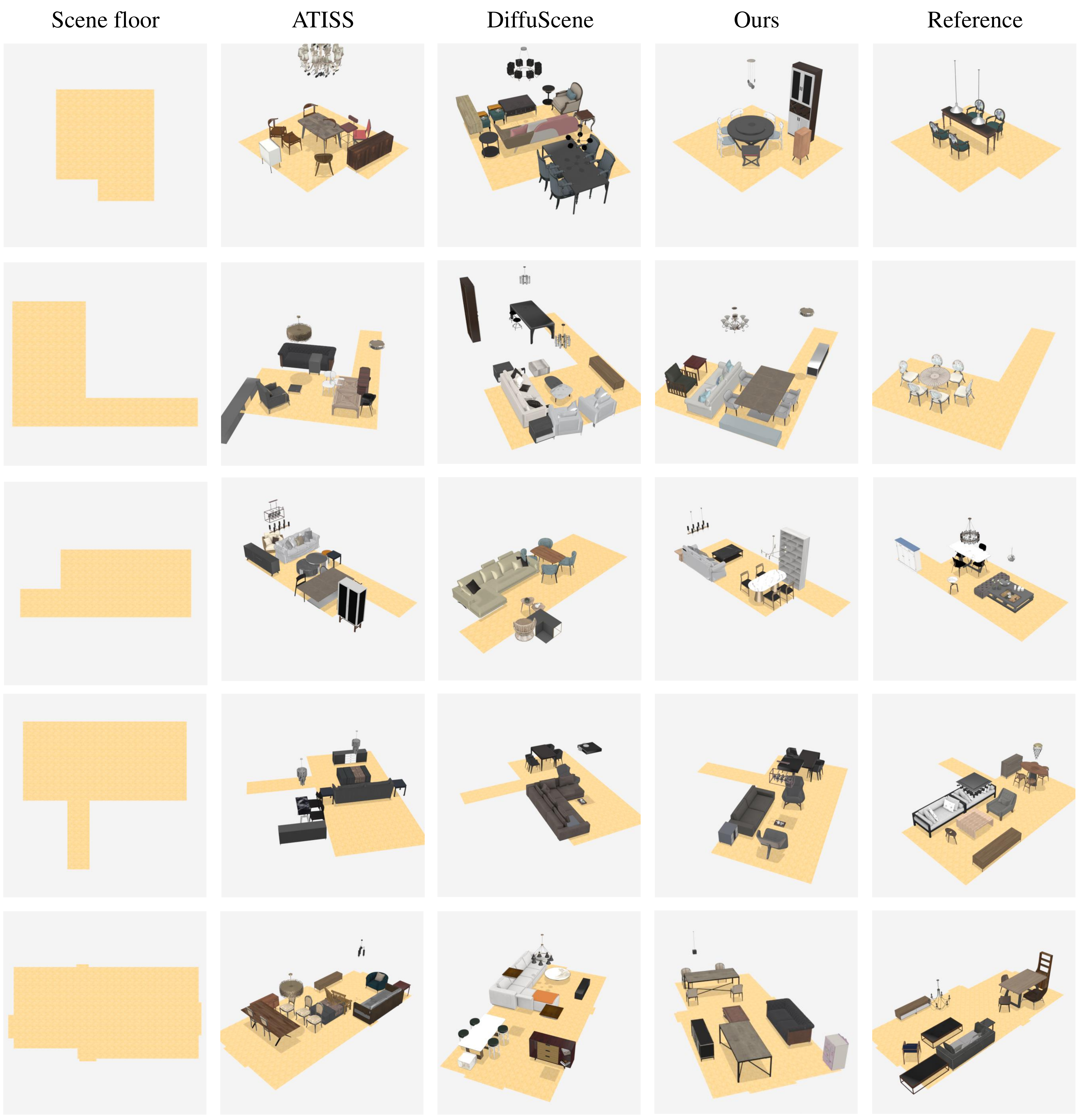}
\caption{Additional results of \textit{living room / dining room} scene synthesis with various floor planes demonstrate that our method can fit well to the given floor layouts.
}
\label{fig:add_cmp3}
\end{figure*}

We present additional results demonstrating the capabilities of our model in various tasks. As shown in ~\Cref{fig:scene_comple}, our model excels in scene completion tasks. \Cref{fig:output diversity} shows our model exhibits randomness and diversity. Furthermore, we tested our model on customized floor plans, achieving strong performance on these tailored designs, as illustrated in~\Cref{fig:more_floor}. Finally, our model's effectiveness in synthesis tasks is showcased in ~\Cref{fig:add_cmp1,fig:add_cmp2,fig:add_cmp3}.

\end{document}